# Data storytelling meets interpretable machine learning: Decoding AI decisions for non-experts without revealing sensitive data and model details

Lemen Chao [a, b], Zixuan Yang [b], Anran Fang [b], Mingran Sun [b], Ming Lei [b, *]

[a] *Key Laboratory of Data Engineering and Knowledge Engineering, Ministry of Education, Renmin University of China*
[b] *School of Information Resource Management, Renmin University of China*

**ABSTRACT**
AI-driven automated decision-making requires both predictive performance and interpretability. Recent advances in interpretable machine learning (IML) provide tools for explaining model predictions, but the technical complexity of these explanations may hinder accessibility to non-experts. To address this challenge, this study integrates data storytelling with IML to enhance the explainability of AI-generated decisions for a broader audience. Following the design science research (DSR) paradigm, this study proposes a formal definition of data storytelling in IML, introduces the DIST Pyramid to align data storytelling with IML, and presents the I–P–O Model to describe their interactions. It further develops an architecture to explain AI decisions through distinct "What-if" and "Why-not" event-generation processes. The architecture also employs data desensitization to protect sensitive input data. To validate the approach, a case study is conducted with the Boston Housing dataset, using SHapley Additive exPlanations (SHAP) values and large language models (LLMs) to generate data stories with And-But-Therefore (ABT) structures. An empirical evaluation shows that 76.4% and 74.3% of respondents rated the "What-if" and "Why-not" data stories as more comprehensible, with significantly higher accessibility scores than traditional SHAP visualizations. The paper concludes with the presentation of a narrative interpretation framework that integrates IML and data storytelling, thereby expanding the research scope as well as the practical applicability of AI decision-making.

**Keywords:** Data Storytelling, Interpretable Machine Learning, AI Decisions, Trustworthy AI, Explainable AI, Large Language Model

## 1. Introduction

Applications of AI in domains such as fraud detection, autonomous vehicles, and precision medicine illustrate the growing prevalence of automated decision-making. Nevertheless, these automated decision-making systems raise concerns about transparency and trustworthiness owing to their underlying black-box models (Lipton, 2016; Górski & Ramakrishna, 2021). Furthermore, corporations often withhold information about their own proprietary AI systems from the public to maximize their profits derived from these systems (Rudin, 2019), thereby making the AI decisions even more opaque. The insufficient understanding of decisions automatically made by machines leads to a lack of trust in automated systems, particularly among individuals without specialized knowledge (Hoffman et al., 2009). It is further exacerbated when these decisions involve critical domains such as medical diagnosis (Caruana et al., 2015). Justified trust in AI models is essential to enabling users to place confidence in their decisions (Hoffman et al., 2018a). This underscores the importance of interpreting complex AI decisions. The European Commission has listed four key ethical principles for trustworthy AI, with explicability identified as one of them (European Commission, 2019). Policies and regulations, such as the EU AI Act (European Commission, 2024), are being implemented, further reinforcing the societal relevance of trustworthy AI. In practical applications, explanations of automated AI decisions must protect trade secrets (Carvalho et al., 2019), data security (Hamon et al., 2022), and personal privacy (Awosika et al., 2024).

Multiple Interpretable Machine Learning (IML) methods, including LIME (Ribeiro, et al., 2016), Anchors (Ribeiro, et al., 2018), and SHAP (Lundberg & Lee, 2017) have been developed to enhance model transparency and trustworthiness. These algorithms facilitate the explanation of an AI model's decision-making process without exposing its internal structure. IML methods are extensively utilized in explainable artificial intelligence (XAI). For instance, Bataineh et al. (2024) designed a novel framework based on XAI and Shapley values for detecting poisoning attacks in collaborative Intrusion Detection Systems used in vehicular networks. Bibi, et al. (2025) reviewed the key XAI techniques, namely LRP and DeepLIFT, applied in the diagnosis of brain diseases.

* Corresponding author: Ming Lei, School of Information Resource Management, Renmin University of China, 59 Zhongguancun Street, Haidian District, Beijing, 100872, China.
*E-mail address*: leimingnick@ruc.edu.cn (M. Lei).

While IML methods are designed for experts, their interpretability remains limited for non-experts. Individuals lacking digital literacy and technical experience often fail to interpret data and understand patterns when complex information is presented Clegg et al., 2020). Non-experts frequently encounter information overload (Chen et al., 2012), and merely visualizing data is inadequate for confident interpretation of the findings (Shao et al., 2024). As a result, researchers have identified key criteria for effective interpretation, such as clarity, transparency, and user comprehensibility (Hoffman et al., 2018b). Systems designed to interpret AI decisions require further improvement. To enhance comprehensibility for non-experts, their design should be interactive, provide accessible explanations, and protect private information (Ali et al., 2023).

Data storytelling presents a promising approach for explaining AI decisions. It involves using data, visualizations, and narratives to convey complex information, insights, or findings in a clear and compelling manner (Dykes, 2019; Martinez-Maldonado, 2023). This approach is particularly effective in enhancing the understanding of AI decisions among non-experts (Kosara & Mackinlay, 2013; Gómez et al., 2023). Researchers have identified multiple advantages of data storytelling compared with traditional ways of presenting data to non-experts, both in terms of delivery and reception. (1) In terms of delivery, data storytelling communicates messages more accurately and effectively by using a richer and more diverse medium of presentation (Segel & Heer, 2010; Zdanovic et al., 2022; Shao et al., 2024). Additionally, combining narratives with visualizations can emphasize key information and improve message delivery efficiency (Ryan, 2016; Echeverria et al., 2018). (2) Regarding reception, data stories enhance users' understanding, evoke emotion and improve retention. A data story enhances understanding and retention, because the incorporated visualizations and narrative elements align with humans' natural aptitude for visual perception, making the presentation more engaging (Knaflic, 2015; Echeverria et al., 2018). Shao et al. (2024) argue that data storytelling can enable viewers to simultaneously understand multiple data points. Furthermore, Data storytelling emotionally engages viewers, thereby facilitating decision-making (Knaflic,2015). This emotional connection is crucial for building trust in AI decisions. Graeber et al. (2024) demonstrated that stories are more easily remembered and recalled than raw statistics, giving them a more enduring influence on viewers' beliefs over time. With these traits and advantages, data storytelling is a desirable tool for explaining AI decisions to non-experts

Therefore, to fill the gap between the interpretability limitations of high-performance AI models and the explanatory needs of non-experts, this study aims to establish a theoretical foundation and implementation workflow based on data storytelling, integrating IML techniques to decode AI decisions for non-experts without revealing sensitive data and model details. This study proposes the following specific objectives:

- Define a hybrid approach that integrates data storytelling with iML for AI explanations, considering model and privacy security, and elucidate the hierarchical progression from prediction model output to non-expert trust.
- Design a general process and an operational workflow to automate the generation of data stories, while embedding model and privacy protection features.
- Implement the data story generation based on the reference workflow and evaluate the explanation results by user-centric metrics, comparing their effectiveness in conveying explanation events.

In addressing these objectives, we develop a set of theoretical and methodological frameworks for data storytelling, integrated with IML methods and grounded in the Design Science Research paradigm, to decode AI decisions for non-experts, protecting sensitive data and model details. The feasibility of these proposals is further validated by automatically generating data stories. This study makes five primary contributions:

- Formally define data storytelling integrated with iML methods for explaining AI decisions without revealing sensitive data and proprietary model details. This formal definition illustrates data storytelling's capacity to support interpretability, transparency, privacy protection, and comprehensibility.
- Propose a Data Storytelling Conceptual Pyramid (DIST Pyramid) to explain AI decisions for non-experts via data storytelling. This framework outlines a hierarchical progression from AI decision generation to non-expert trust, mediated through interpretation and data storytelling.
- Propose a Data Storytelling Interactive Process to incorporate analytical models, story models, and narrative models. This interactive process further expands the interactive function of data storytelling to better communicate with non-experts.
- Develop an operable and flexible data storytelling workflow, the Data Storytelling Reference workflow, designed to convert AI decisions into structured narratives.
- Conduct a case study that employs the representative ABT and other story structures as well as iML methods to explain the what-if and why-not data story events for non-experts. Our demonstration utilizes advanced LLMs to generate the text narratives of data stories, facilitated by visualizations of SHAP plots. The evaluation demonstrated the overall positive impact of our ABT data stories.

The paper is organized as follows. Section 2 reviews the background, focusing on data storytelling concepts, techniques, and emerging trends. Section 3 describes the methodologies employed in this study. Section 4 presents our main theoretical and methodological contributions organized into four key areas. Section 5 provides a case study and evaluates ABT data stories and their conversion into alternative story structures. Section 6 discusses the implications and limitations of data storytelling. Finally, Section 7 concludes with directions for future research.

## 2. Background and related work

### *2.1. What is data storytelling*

Storytelling, one of the oldest and most powerful means of interpretation, has evolved through four distinct phases of development: tribal narrative, static narrative, post-linear narrative, and global narrative (McDowell, 2019). Its strength lies in its cognitive accessibility, facilitating both comprehension and retention. A comparative study at Stanford University shows that 63% of respondents could recall narratives, while only 5% remembered statistics (Small, et al., 2007). Storytelling is not only easily comprehensible but also highly effective in persuading audiences. Research from Carnegie Mellon University demonstrates that storytelling raised donations more than double those of fact-based solicitation (Heath & Heath, 2007). Moreover, an analysis of 500 TED Talks conducted by Harvard University reveals that narratives accounted for 65% of the content (Gallo, 2019).

Data storytelling is the ability to effectively communicate insights from a dataset through narratives and visualizations. It contextualizes data insights and inspires action from audiences (Catherine, 2021). Dykes (2019) identifies three fundamental elements of data storytelling (data, visualization, and narrative), each playing a distinct yet complementary role in conveying decisions. These elements collectively serve three primary functions: explain, enlighten, and engage. The explain function uses narrative techniques to clarify data and its underlying patterns. The enlighten function integrates data and visualization to enhance understanding and reveal insights. The engage function combines narrative and visualization to capture attention and encourage active participation. These functions work synergistically to explain decision-making to the target audience and drive desired actions. Data storytelling encompasses a range of interconnected concepts, each emphasizing distinct aspects of the storytelling process, and can be considered the convergence of all these concepts (Table 1).

**Table 1**
Concepts related to data storytelling.

| Concept | Description | Reference |
|---|---|---|
| Data-driven storytelling | Focuses on data, with narratives either grounded in or supported by data visualizations | Riche et al., 2018 |
| Visual storytelling | Utilizes visualizations to convey information, but does not fully address the complexities of trust in AI | Beauxis-Aussalet et al., 2021; Williams, 2019; Hudon et al., 2021 |
| Analytical storytelling | Organizes data-driven insights into a structured narrative, directing the audience's focus toward key data points | Gagnon and Caya, 2020 |
| Interactive storytelling | Promotes audience engagement, enabling dynamic, nonlinear, and interactive narration | Park, 2017 |
| Storytelling with data | Emphasizes data as the foundation for storytelling, distinguishing it from conventional storytelling by relying on data evidence to support the narrative | Knaflic, 2019 |
| Digital storytelling | Leverages digital media to present stories, as opposed to traditional media (e.g., print, radio, face-to-face storytelling) | Miller, 2019 |

### *2.2. Data storytelling versus traditional storytelling*

Data storytelling and traditional storytelling vary in six key aspects, as outlined in Table 2. Data storytelling is typically driven by specific business objectives and relies extensively on real business data or algorithms. In contrast, traditional storytelling primarily serves broader purposes, such as entertainment or education, often involving fictional narratives or imaginary elements. Data storytelling is algorithmically generated, whereas traditional storytelling is manually crafted. Additionally, data stories typically follow a non-linear, audience-driven narrative, while traditional stories are linear and author-driven. Data storytelling also differs from traditional storytelling in its theoretical foundations: the former is rooted in fields such as data science and cognitive science, while the latter is based on literature, philosophy, and folklore. Lastly, data stories and traditional stories have different lifespans. Data stories typically have short-term relevance due to their business-oriented nature, whereas traditional stories persist over time by exploring universal themes that transcend specific business concerns.

**Table 2**
Key differences between data storytelling and traditional storytelling.

| Aspect | Data storytelling | Traditional storytelling |
|---|---|---|
| Motivations | Focused on specific business objectives | Aimed at broader purposes (e.g., entertainment, education) |
| Content | Real business data or algorithms | Fictional or imaginative content |
| Methods | Algorithmically generated | Manually crafted |
| Models | Non-linear model, audience-driven | Linear model, author-driven |
| Theoretical foundations | Data science, big data analysis, infographics, and cognitive science | Literature, religion, philosophy, and folklore |
| Lifespan | Short-term | Long-term |

### *2.3. Data storytelling versus data visualization*

Theoretically, data storytelling and data visualization represent distinct approaches (Fig. 1). Data visualization addresses the perception of data, primarily focusing on expressing the meaning of data. Data storytelling further addresses the cognition of data, embedding it in a narrative to enhance understanding by engaging broader cognitive processes. While visualization focuses on data perception through a visual system, storytelling expands beyond this single channel by incorporating multiple or mixed channels, such as hearing, touch, and taste, to foster a more comprehensive understanding of the data. Storytelling Alice, for instance, engages middle-school girls in computer programming by allowing them to create animated movies (Kelleher & Pausch, 2007).

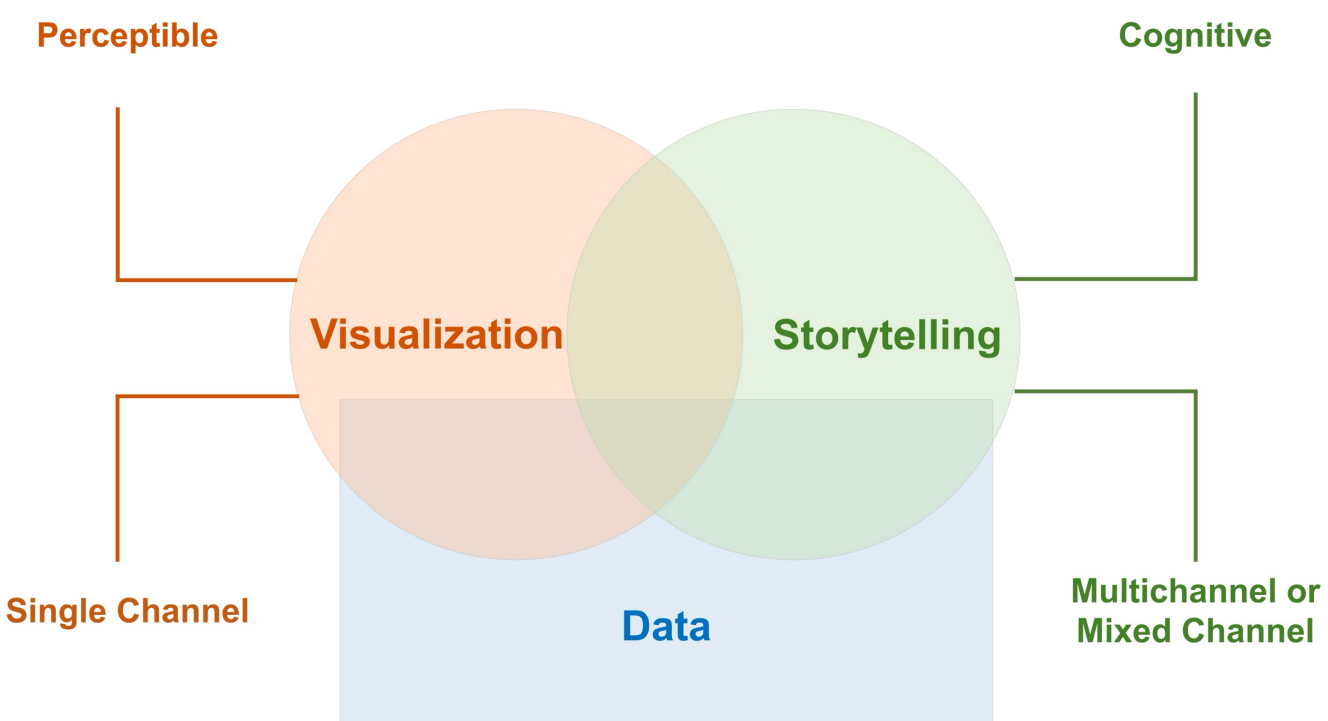

**Fig. 1.** Comparison of data storytelling and data visualization approaches.

However, in practice, data visualization is used as a means of data storytelling, enhancing both the perceptibility and interpretability impact of the narrative. Visual storytelling is a prevailing approach in data storytelling and has become a widely adopted method in early data storytelling tools. In tools such as Tableau and PowerBI, data storytelling capabilities are primarily implemented via visual storytelling techniques. In the early stages of data storytelling, visualization was prioritized due to its relatively straightforward implementation. By contrast, integrating text, voice, or image-based technologies posed greater technical complexities. Nonetheless, with the rise of Large Multimodal Models (LMMs), visualization may not be regarded as the predominant method in data storytelling. Instead, a broader range of modalities and hybrid forms has the potential to promote the next generation of data storytelling.

### *2.4. Data storytelling frameworks and techniques*

Various frameworks and techniques for data storytelling have been proposed, ranging from narrative structures and models for story generation to practical tools for implementation. Davenport (2014) proposed several types of data stories, such as reporting and Eureka stories, which were classified along the dimensions of time, focus, depth, and method. Segel and Heer (2010) presented three schemas balancing author-driven and reader-driven narratives: Martini Glass, Interactive Slide Show, and Drill-Down Story. Multiple tools or techniques have been developed to generate data stories. Li et al. (2024) conducted a systematic review of data storytelling tools from the perspective of human-AI collaboration, examining the respective roles of humans and AI at different stages of data storytelling. For visual narratives, automatic chart generation tools such as SAGE (Roth et al., 1994) and WebFOCUS (Webfocusinfocenter, 2025), annotation tools like ChartAccent (Ren et al., 2017), and data visualization tools like Tableau are utilized. For text narratives, natural language generation (Mellish et al., 2006) techniques have advanced significantly in recent years. Trim et al. (2024) designed a writing unit centered on data exploration and storytelling, with the objective of enabling students to understand how biases in datasets can lead to errors in AI models' decision-making processes. For multimodal narratives, data story products that leverage multimedia techniques, especially virtual reality (VR) and augmented reality (AR), are increasingly employed to convey data-driven insights. Ferracani et al. (2024) implemented an application based on GPT-4 and an image generation pipeline enabling tourists to generate personalized city tour stories. Jain et al. (2024) explored the implementation of various forms of immersive data storytelling, including "360 video" and "Gamification" and proposed a foundational design framework for these emerging forms.

### *2.5. Data storytelling for XAI*

There is a growing trend toward integrating iML and XAI techniques with data storytelling to produce human-friendly explanations. In practice, combining iML's feature attribution with data storytelling often involves template-based or LLM-based generation. LLMs offer a promising approach to XAI by transforming complex model outputs into easy-to-understand narratives, bridging the gap between model decisions and human interpretability.

XAINES projects (Hartmann et al., 2022) distinguish two types of AI explanation narratives: "ML narratives", which describe the model-internal causal reasoning, and "domain narratives", which describe the outcome in terms of real-world events and domain context. Martens et al. (2025) introduced the concept of "XAIstories", which leverage LLMs to generate user-oriented narratives from technical explanations. In their framework, outputs from XAI methods (such as SHAP and counterfactual explanations) are fed into an LLM (GPT-4) prompt to produce a natural language story explaining a given model prediction. Bilal et al. (2025) demonstrated that integrating visual and textual explanations enhances the understanding of models' underlying reasoning process, both visually and contextually.

Furthermore, recent research has increasingly tailored data storytelling approaches to specific domains. The explanations are formulated with domain context, jargon, and narrative style appropriate to the field. For example, within the medical domain, the storytelling XAI framework (Dubey et al., 2024) for medical imaging aims to generate comprehensive visual and textual reports. It integrates visualization results derived from IML techniques with textual narrative interpretations for radiologists, thereby justifying the model findings presented.

These findings highlight the synthesis of data storytelling with XAI as a means of making AI decisions more comprehensible, accessible, and convincing to broad audiences. However, existing studies have not yet proposed a general-purpose theoretical framework or an implementation workflow for applying data storytelling to explain AI decisions.

## 3. Methodology

### *3.1. Design Science Research*

Design Science Research (Hevner et al., 2004; Vom Brocke et al., 2020) is a problem-solving paradigm grounded in engineering and the artificial sciences. It aims to advance human and organizational capabilities by creating innovative artifacts, such as models, methods, and frameworks, to solve real-world problems. This study adopts Design Science Research (DSR) because of its advantages in creating and iteratively refining artifacts. DSR is particularly suitable for addressing complex, real-world challenges, such as utilizing data storytelling to decode AI decisions for non-experts. The steps for applying the DSR methodology in this study are as follows:

1) Problem identification and motivation: Define the challenge of rendering AI decisions interpretable to non-experts while protecting sensitive data and proprietary model details from disclosure.
2) Definition of objectives for a solution: Establish the objective of designing an artifact that clearly communicates AI decisions by integrating iML methods with storytelling. This objective should be achieved while ensuring the confidentiality of sensitive data.
3) Design and development: Develop an artifact that combines iML techniques, such as LIME, Anchors, and SHAP, with data storytelling and privacy-preserving methods (e.g., differential privacy) to provide understandable explanations without compromising sensitive information.
4) Demonstration: Implement the artifact in real-world scenarios to help decode AI decision-making in a way grounded in faithfulness to data objectivity, while improving comprehensibility for non-experts.
5) Evaluation: Assess the artifact's effectiveness in providing interpretable explanations to non-experts and ensuring the protection of sensitive data.
6) Communication: Present the research findings to the testing team, ensuring that the explanations are accessible, actionable, and contextually relevant, while preserving data privacy.

This framework combines explainability and privacy-enhancing technologies to ensure that AI decisions are transparent, interpretable, and understandable to non-experts without exposing sensitive data or model details.

*3.2. ABT storytelling framework and its mapping to other story structures*

Considering the ability to automatically adapt to varying narrative needs, the selection of an appropriate story structure is a critical step in the data storytelling process. In the field of data storytelling, a variety of story structures are widely applied, including the ABT Story (Olson, 2015), Freytag's Pyramid (Yang et al., 2021), Hero's Journey (Mahoney & Nickerson, 2022), the Boy Meets Girl (Vonnegut, 2007), Three-Act Structure (Field, 2005), The Golden Circle (Sinek, 2009), STAR Method (MIT CAPD, 2025), SCQA Framework (Minto, 1996), SPSN Framework (Zawadzki, 2018), 5W1H Framework (Li et al., 2023), CAR Framework (Spodek, 2020), and AIDA Model (Hassan, et al., 2015).

The objective of storytelling in this study is to provide multiple story structure options for the same data and black-box model, which requires the flexibility to switch between them automatically. To this end, the ABT story structure is established as the fundamental storytelling framework, due to its significant advantages in explaining model decisions:

- Simplicity: The logic behind this linear story structure is simple and straightforward. It turns the abstract and complex interpretation of model decisions into a concise data story that is easy to understand, even for non-experts.
- Integrity: An ABT story has a complete causal chain. The And part sets the background for the story to happen; the But part raises a problem or conflict based on the background setting and drives the plot forward; and the Therefore part solves the problem, reaching a balance and forming a narrative closure. After being guided through the whole process, non-experts are supposed to gain a deep insight into the model's decision.
- Applicability: The structure of an ABT story is applicable in various contexts, and its logic is the basis of many other story types, which makes it possible to map an ABT story to almost all other types of stories.

The storytelling framework mapping methodology is outlined in Table 3: the story is initially constructed using the ABT story structure and then mapped to other story structures according to the user's demands.

**Table 3**
The mapping relationship between the ABT storytelling framework and other storytelling frameworks.

| Data storytelling framework | A: And | B: But | T: Therefore |
|---|---|---|---|
| ABT Story | Setting the background, describing the current state | Presenting the problem, conflict, or challenge | Providing the solution or action |
| Freytag's Pyramid | Exposition: Background setting, introduction of characters and environment | Rising Action + Climax: Problem/conflict intensifies, climax emerges | Falling Action + Denouement: Problem resolution, conclusion |
| Hero's Journey | Call to Adventure + Refusal of the Call: The hero's invitation and initial hesitation | Trials + Climax: Challenges and conflict reaches its peak | Return with the Elixir: Problem resolution, returning with gained wisdom |
| Boy Meets Girl | Starts in a positive state, forms a meaningful connection | Faces loss, conflict, or dilemma requiring a decision | Resolves the conflict, finds resolution, and achieves happiness |
| Three-Act Structure | Act 1: Setup: Establishing background and context | Act 2: Confrontation: Conflict escalation and challenge | Act 3: Resolution: Conflict resolution, deriving conclusion |
| Golden Circle | Why: The mission or vision of the organization or product | How: The methods or ways to address the problem | What: Providing solutions, demonstrating outcomes or actions |

| Data storytelling framework | A: And | B: But | T: Therefore |
|---|---|---|---|
| STAR Method (Situation-Task-Action-Result) | Situation: Describing the current scenario or context | Task + Action: Challenges and steps taken to address them | Result: Presenting the outcomes or impacts of the solution |
| SCQA Framework (Situation-Complication-Question-Answer) | Situation: Establishing the background and context | Complication: Introducing the problem or conflict | Question + Answer: Addressing the problem by posing and answering a key question |
| SPSN Framework (Situation-Problem-Solution-Next Steps) | Situation: Describing the current state and context | Problem: Defining the existence of the problem or challenge | Solution: Proposing a resolution and showcasing results |
| 5W1H Framework | Who/What/Where: Describing the problem's background | Why: Explaining why the issue exists, exploring reasons | How: Proposing solutions and next actions to address the issue |
| CAR Framework (Context-Action-Result) | Context: Describing the background and current state | Action: Measures or actions taken | Result: Outcomes of the actions taken |
| AIDA Model (Attention-Interest-Desire-Action) | Attention: Capturing the audience's attention | Interest: Generating interest in the topic | Desire + Action: Stimulating the audience's desire to act |

This mapping approach ensures the creation of clear and contextually relevant explanations, while facilitating the flexible adaptation of various storytelling structures, thereby improving the interpretability of AI decision-making processes.

## 4. Integration of data storytelling with IML

### *4.1. Formalized description of data storytelling*

Both theoretical discussions and IT implementations of data storytelling necessitate a formalized definition that precisely describes how to integrate black-box models, sensitive data, interpretability techniques, and data storytelling modeling to achieve data storytelling. In our study, we define data storytelling as follows:

$$F_{story}(f, x_s) = S\left(C\left(I_{exp}(f, D(x_s)), A(x_s)\right), T_{story}\right) \tag{4.1}$$

- $f$ (Black-box model): a black-box model that generates specific decisions, typically serves as an example of complex, non-interpretable models (e.g., deep neural networks, large language models).
- $x_s$ (Sensitive Data): the original input data containing sensitive or confidential information that requires protection.
- $D(x_s)$ (Data desensitization function): the function that applies desensitization techniques to the sensitive data $x_s$, such as differential privacy ($\varepsilon$-$\delta$ differential privacy), secure multi-party computation (SMPC), federated learning, data perturbation, generalization, and data masking, to protect sensitive information.
- $I_{exp}(f, D(x_s))$ (iML method): an interpretation method (e.g., LIME, SHAP, Anchors) applied to desensitized data to derive insights from a decision made by black-box model $f$.
- $A(x_s)$ (Data augmentation function): a function that generates supplementary data beyond desensitized data to enrich data storytelling. It constructs essential narrative elements such as characters, events, and plots, ensuring the generated insights are engaging and contextually relevant.

- $C\left(I_{exp}\left(f, D(x_s)\right), A(x_s)\right)$(Contextualization function): combines the explanation $I_{exp}\left(f, D(x_s)\right)$ with the augmented data $A(x_s)$, enriching the explanation with real-world context to make it relevant and actionable.
- $S\left(\cdot, T_{story}\right)$ (Structured storytelling function): this function transforms the contextually enriched explanation into a structured narrative, guided by a chosen storytelling framework $T_{story}$ (e.g., ABT, Pyramid Hero's Journey), ensuring the explanation is clear and engaging for non-expert users.

The process begins by applying $I_{exp}\left(f, D(x_s)\right)$, generating an interpretable explanation of the black-box model's decision based on original inputs. Then, $A(x_s)$ produces a desensitized version of the input data, protecting privacy while preserving the relevant features necessary for explanation. The explanation is further contextualized by combining it with $A(x_s)$, making it applicable to specific real-world scenarios. Finally, the structured storytelling function $S()$ organizes and presents the enriched explanation using a storytelling framework $T_{story}$, ultimately producing compelling stories that are easily understandable for non-experts.

This formalized definition presents a systematic methodology for interpreting decisions of black-box models while protecting sensitive data. Combining privacy-preserving techniques with iML methods and structured storytelling, this methodology enables complex model decisions to be transparent, interpretable, and understandable to non-experts without compromising privacy.

*4.2. Data Storytelling Conceptual Pyramid*

The DIST (Decision-Interpretation-Storytelling-Trust) Pyramid provides a hierarchical conceptual framework for explaining AI decisions for non-experts. It describes a hierarchical progression from interpretability through transparency and ultimately to trustworthiness, demonstrating how each stage systematically builds upon and reinforces the previous one. The design of the DIST Pyramid in Fig. 2 is grounded in Section 4.1, with a primary concern for elucidating the four essential stages that bridge decision-making and trust.

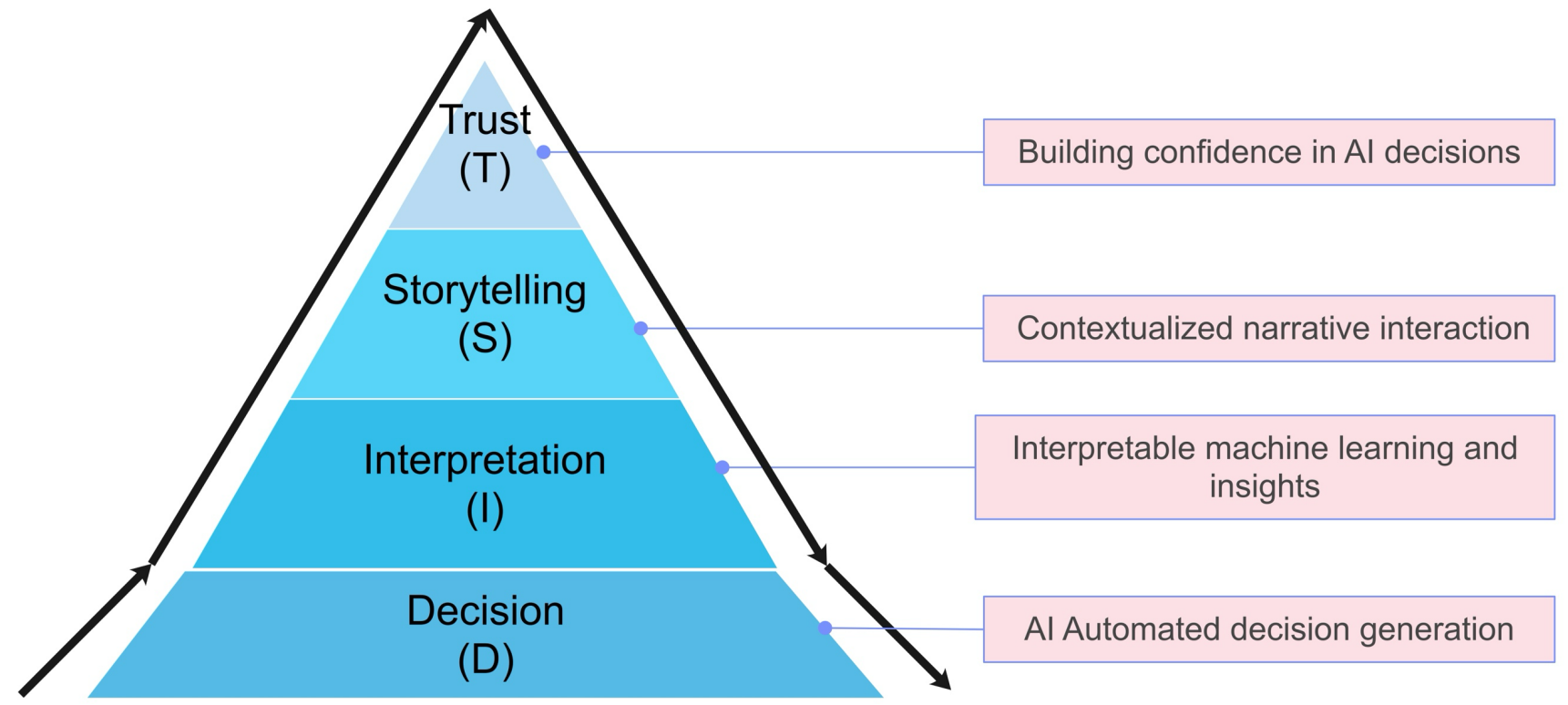

**Fig. 2.** The DIST (Decision-Interpretation-Storytelling-Trust) Pyramid.

1) The decision layer represents decisions derived from black-box models. It forms the foundation of the pyramid, providing the decisions that will be interpreted and communicated in the upper layers. Typically, these decisions are often difficult to convey to non-experts in an interpretable manner, which is why interpretation methods and storytelling techniques are introduced and integrated.
2) The interpretation layer applies explainability methods, such as LIME, SHAP, and Anchors, to model outputs, providing transparent insights while preserving privacy. This layer clarifies how the AI generates its decisions, making the complex model outputs more understandable. Interpretation forms the bridge between the AI decisions and their meaningful storytelling to users.
3) The storytelling layer converts the interpretation results into a narrative that is both relatable and actionable for audiences. A well-structured explanation enhances the relevance and accessibility of the decision-making process, ensuring the interpretation resonates with non-expert audiences.
4) The trust layer is located at the top of the pyramid. Trust is the outcome established through transparency and relevance. This layer ensures that AI decisions are not only interpretable but also contextually relevant to users. Trust is built when users gain confidence in the AI system through storytelling explanations that are both comprehensible and aligned with their context.

### 4.3. Data Storytelling Interactive Process

The Data Storytelling Interactive Process, based on Borjigin (2021), is introduced to provide theoretical and methodological support for effectively communicating AI decisions to non-experts. It consists of three sub-models—the analytic model (I-model), the story model (P-model), and the narrative model (O-model), as illustrated in Fig. 3. These sub-models represent the essential stages of data storytelling: data analysis, story construction, and audience engagement. The integration of these stages enables complex AI decision-making processes to be explained transparently and meaningfully while preserving privacy.

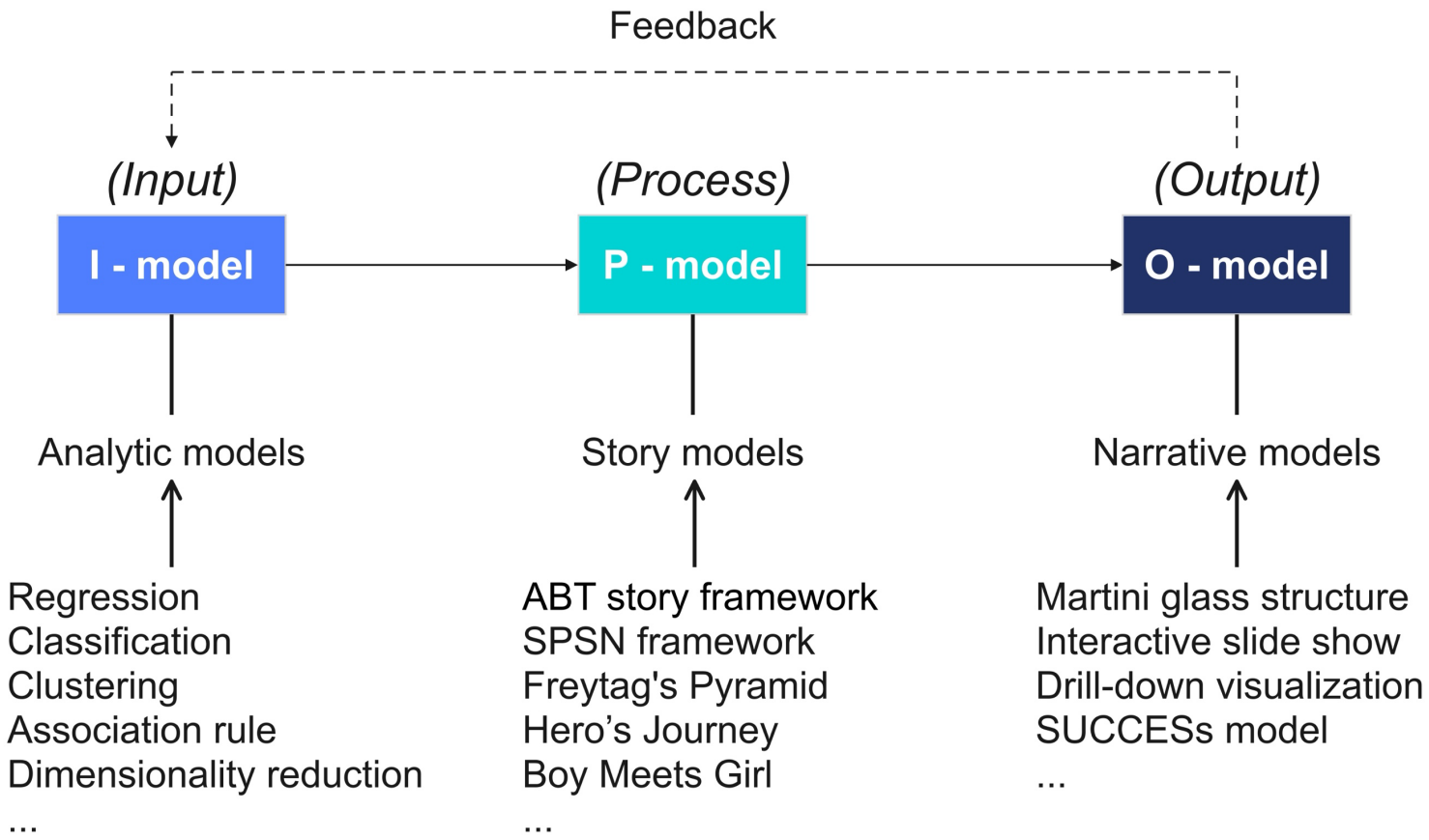

**Fig. 3.** Data Storytelling Interactive Process.

- The I-model: The input model (or the analytical model) generates the input data or potential events for the data story. This model serves as the foundation of data storytelling. Common analytic models include regression, classification, clustering, association rule analysis, and dimensionality reduction.
- The P-model: The process model (or the story model) integrates the core elements of a data story and defines their interconnections, serving as a bridge between the analytical and narrative models. Storytelling structures refer to those discussed in Section 3.2, including the ABT story framework, the SPSN framework, Freytag's Pyramid, Hero's Journey, and Boy Meets Girl. These frameworks help structure key components of the story, such as the situation, problem, solution, and narrative arcs.
- The O-model: The output model (or the narrative model) is responsible for presenting the story to the target audience. It chooses an appropriate story model and presents its outcomes in a narrative form tailored to the needs of the target audience. The recommended narrative models for data storytelling include the Martini Glass Structure (Weber et al., 2018), Interactive Slide Show (Segel & Heer, 2010), Drill-Down Visualization (Segel & Heer, 2010), and the SUCCESs Model (Finkler & León, 2019). The incorporation of these models enhances the effectiveness of data storytelling by improving audiences' interaction and fostering emotional engagement.

The interactions among these models are bidirectional, as the output from the O-models can provide feedback that influences the input to the I-models, thereby driving model refinement. Similarly, the insights generated by the P-models trigger iterative adjustments or modifications to the initial data processed by the I-models.

### 4.4. Data Storytelling Reference Workflow

This section further introduces a reference workflow for data storytelling applications as a complement to the practical implementation of the Data Storytelling Interactive Process, particularly bridging the gap between the analytical model and the story model by incorporating the iML methods. The workflow consists of five key modules (Fig. 4), each of which implements one of the core functions necessary for converting a black-box model decision into accessible data stories:

1) The **Predictor** represents an AI prediction system that uses a black-box model (M) to process the user's submitted dataset (D) and generate a prediction result (R). This module is responsible for producing the decision that will be further interpreted for non-experts. In most cases, this module functions as a business system and only needs to expose an API (Application Programming Interface) for the data storytelling system, without the necessity for redevelopment.
2) The **Interpreter** trains a surrogate model to simulate the input/output behavior of the original model (M). It samples and perturbs

the original dataset (D), applying data desensitization techniques to transform it into a desensitized dataset (D'). Based on this, the module employs iML techniques (e.g., LIME, SHAP, Anchors) to provide a technical-level explanation of the desensitized dataset (D'), thereby producing a surrogate model (M').

3) The **Modeler** defines the data story model (DSM) with respect to the insights derived from the surrogate model (M'). The story model selects a specific storytelling structure, such as ABT storytelling framework or Freytag's Pyramid, and organizes the essential elements and relationships within a data story.
4) The **Scripter** converts the story model into a script (S), which is refined using a script debugger. Another dataset (D'') can be used as input to the surrogate model (M'), resulting in a prediction R''. This exploratory analysis process could generate components of the story content, such as story events. Formal description script S of the data story is then automatically tested and validated by the domain ontology or business rule system (O). This ensures that the data story is accurate and ready for narration in a format that can be interpreted by both machines and humans.
5) The **Narrator** delivers the final story script (S) to non-experts using specific storytelling methods, such as natural language generation, visualizations, films, interactive media, and augmented reality. This sub-module focuses on the final step of data storytelling, namely narrating the story to non-experts.

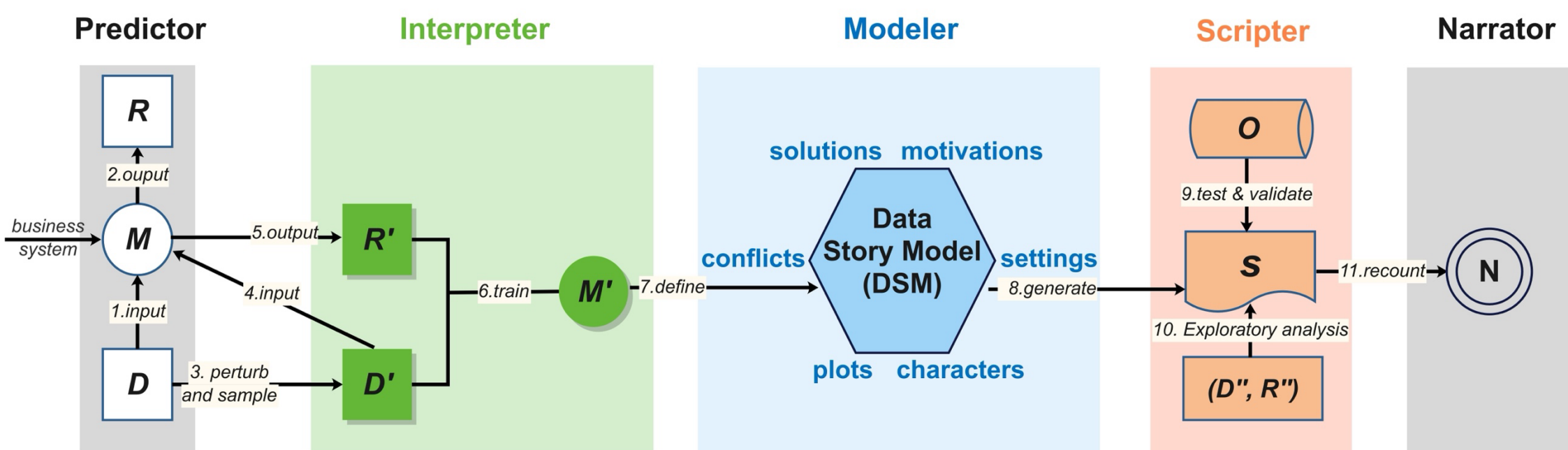

**Fig. 4.** Data Storytelling Reference Workflow.

The automatic generation of candidate events is a critical step in automatic data storytelling, as events are fundamental building blocks of a story plot. The number of events, their significance, and their sequence are all key factors in determining the quality of the story plot. As discussed in Section 3.2, structuring events into a coherent plot is essential for crafting engaging stories. However, not all candidate events are included in the plot: some need to be filtered, selected, or sampled before being organized into a storytelling structure. The selection and structuring of these events into a coherent plot enable the reference framework to translate black-box model decisions into clear, actionable data stories without exposing confidential information. This study identifies two types of event generation and selection in data storytelling: what-if events and why-not events (Fig. 5).

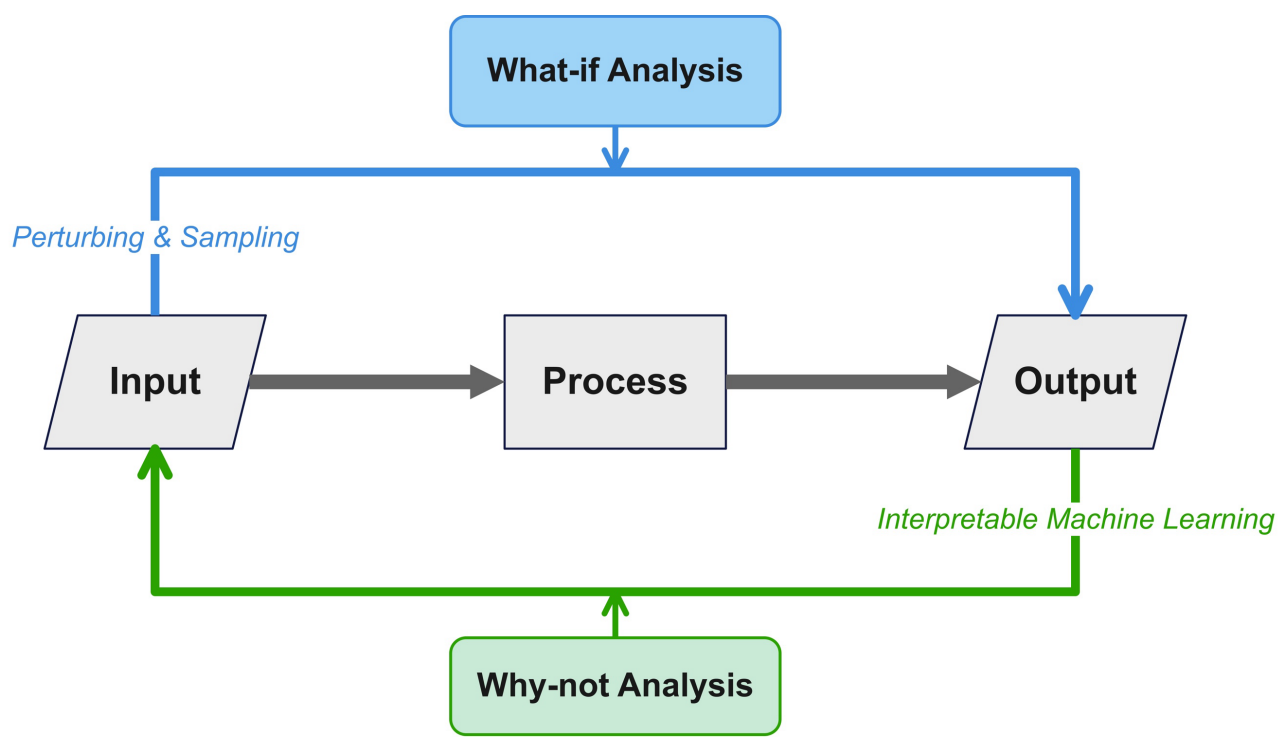

**Fig. 5.** Approaches to generating candidate events for data storytelling.

- What-if analysis is commonly used to perturb and sample the input data of the real model (I-Model), enabling the observation of how such alterations affect model predictions. Essentially, What-if analysis involves counterfactual explanations. We refer to the story events generated through What-if analysis as What-if events. In data storytelling, What-if analysis is a narrative technique that uses explanatory variables and can be used to generate story content grounded in correlation analysis.

- Why-not analysis typically involves modifying the output data of the real model (I-Model) to examine the underlying factors influencing the outputs. The events generated through Why-not analysis are referred to as Why-not events. In the context of data storytelling, Why-not analysis serves as a narrative technique that relies on explained variables rather than explanatory variables, and it can be applied to generate story content based on causal inference.

## 5. Case study: Explaining black-box AI decisions with ABT data stories

In this section, we present a comprehensive case study that demonstrates the application of an IML algorithm to explain a prediction generated by a black-box model. This case study leverages the approaches for integrating data storytelling with Interpretable Machine Learning, as proposed in Section 4, with a specific focus on the automatic generation of what-if and why-not events. We use SHAP to generate detailed explanations for the XGBoost model trained on the Boston Housing dataset (Harrison & Rubinfeld, 1978). These explanations are conveyed through the ABT (And, But, and Therefore) storytelling framework, enabling effective communication of model outputs to a non-expert audience, thereby enhancing transparency and trustworthiness in AI-driven decision-making.

The generated ABT data stories are mapped onto alternative narrative structures, including Freytag's Pyramid, the CAR framework, the SPSN framework, and the STAR method, to facilitate the communication of insights through varied storytelling approaches. Our privacy protection strategy consists of two integral components. The first component introduces noise during the model training phase, ensuring the generation of reasonable explanations while safeguarding sensitive data. The second component involves restricting the granularity of explanations presented in the data stories, especially omitting model names, specific parameters, and code details. This is achieved through a template-based reorganization of IML outputs. As a result, the automatically generated ABT story presents only the users' input data, prediction value, and corresponding feature contributions without revealing any other data information and model details.

Although this case study specifically employs XGBoost and SHAP to generate explanations for black-box model outputs, the Data Storytelling Architecture is sufficiently flexible and can be adapted to a wide range of predictive models and IML algorithms.

### *5.1. Setup*

#### *5.1.1. Description of the dataset*

We use the Boston Housing dataset, a classic dataset with a rich business background, to demonstrate our automated data storytelling workflow. This dataset contains 506 samples and 14 features that provide information about the characteristics of Boston neighborhoods that can influence housing prices. The key features include:

- rm: average number of rooms per dwelling.
- dis: weighted distances to five Boston employment centers.
- nox: concentration of nitrogen oxides in the air (parts per 10 million).
- tax: property tax rate per $10,000.
- crim: per capita crime rate by town.
- ptratio: pupil-teacher ratio by town.
- lstat: percentage of lower status of the population.

The target variable is "medv", which indicates the median house price for owner-occupied homes in a given area, measured in thousands of USD.

#### *5.1.2. Differential privacy processing*

Although the Boston Housing dataset does not contain sensitive personal information, we employ differential privacy techniques to demonstrate the utility of our proposed framework. In particular, we consider the "tax" feature to be sensitive because fluctuations in taxation rates can potentially reveal underlying financial or socioeconomic information about specific neighborhoods. To mitigate privacy risks while maintaining analytical utility, a Gaussian mechanism (Google, 2020) with $\varepsilon = 5$ is used to add noise to the "tax" feature before training the black-box model.

#### *5.1.3. Models*

We train XGBoost as the black-box model to make predictions of house prices. XGBoost (Chen & Guestrin, 2016) is a scalable tree boosting system based on Gradient Boosting Regression Trees (GBRT) or Gradient Boosting Decision Trees (GBDT) to solve regression or classification problems. XGBoost requires a large number of trees to make predictions, rendering its decision-making process difficult to interpret.

As for iML algorithms, we choose SHAP as the interpreter to explain XGBoost's prediction outputs. SHAP employs concepts from cooperative game theory to compute Shapley values, which provide not only local and global feature contributions but also concise and pleasant visualizations for interpretation.

*5.1.4. LLMs usage*

Given the rapid development of LLMs, the automatic generation of data stories has emerged as an increasingly competitive approach. The foundation for this is an effective data story script, according to which LLMs can generate comprehensive data stories. Therefore, we first develop the why-not and what-if event scripts based on the ABT storytelling framework to ensure the overall data story structure and then rewrite them as prompts to drive LLMs in generating texts grounded in SHAP explanations. The detailed script design will be discussed in Section 5.2.1. In this case study, we choose open-source LLMs, Qwen3, Llama 4, and Deepseek R1 as our data story text generators. Qwen3 possesses world-leading language understanding and generation capabilities, earning it a top spot in the recent LiveBench rankings (White et al., 2025). Llama 4 leverages a mixture-of-experts architecture to offer industry-leading performance in text and image understanding. DeepSeek R1 has significantly improved its depth of reasoning and inference capabilities, and its overall performance rivals the leading models, such as O3 and Gemini 2.5 Pro. In particular, we selected Qwen3-235B-A22B, Llama-4-Scout-17B-16E-Instruct, and Deepseek-R1-05/28 to infer our ABT data stories (Table 4). Qwen3-235B-A22B thinking mode and Deepseek-R1-05/28 are large reasoning models with deep thinking steps, while Qwen3-235B-A22B non-thinking mode and Llama-4-Scout-17B-16E-Instruct are standard instruct-driven LLMs. The temperature parameter is set to 0 to ensure that the generated content strictly follows our data story structure. The entire ABT storytelling framework is inferred step by step, making it particularly well-suited to the Chain-of-Thought strategy (Wei et al., 2022). As a result, we employ a template-based Chain-of-Thought (CoT) design for prompt engineering.

**Table 4**

Comparison of the LLMs used in this case study, including examples of generated "Therefore" part for What-if and Why-not events.

| Model name | Parameters | Thinking type | What-if examples | Why-not examples |
|---|---|---|---|---|
| Qwen3-235B-A22B | 235B | Thinking | Adjusting rm and dis alters the prediction, yet nox emerges as a critical unmodified factor, revealing hidden model dependencies. | The why-not analysis shows the model's prediction primarily depends on rm's negative influence and key factors like lstat and ptratio, explaining why alternative outcomes are statistically less supported. |
| Qwen3-235B-A22B | 235B | Non-thinking | Changing "rm" and "dis" increases the prediction because "rm" strongly boosts value, while "dis" and "nox" adjust contributions in response. | The model's higher prediction mainly stems from key factors like rm, Istat, and ptratio, which together significantly influence the outcome compared to the average. |
| Llama-4-Scout-17B-16E-Instruct | 109B | Non-thinking | The what-if analysis reveals that increasing "rm" and "dis" values leads to a higher predicted median house value, mainly driven by changes in "rm", "dis", and "nox". | The model's prediction of 24.0976 is higher than average due to significant contributions from rm, lstat, and ptratio, explaining the deviation from other possible results. |
| Deepseek-R1-05/28 | 685B | Thinking | Adjusting rm and dis increased medv predictions, with rm, dis, and nox driving this change most significantly. | This why-not analysis resolves your doubt by revealing how the model's prediction emerged from conflicting feature influences. |

*5.2. Automatic generation of ABT data stories*

When the ABT storytelling framework is applied to the interpretation of the black-box model, generally, the "And" part describes the current situation and the basic prediction of the model; the "But" part explains the model's decision, pinpointing key factors and their influence; and the "Therefore" part is the summary of events' consequences and the explanatory conclusion, providing insights to help non-experts make decisions.

*5.2.1. ABT storytelling process design*

After seeing a black-box model's prediction, a non-expert user may have two types of questions: what-if and why-not. The whole design of the ABT story is based on these two questions. We employ three sections in presenting ABT data stories. The first section utilizes a structured template that follows the ABT storytelling framework. The "And" and "But" parts, and the data summary of the "Therefore" part are rigorously designed using a scripted template to ensure objectivity. The second section involves natural language generation, where insights are generated in conclusion of the "Therefore" part using LLMs based on the contextual data provided in the first section. The third section incorporates SHAP data visualization, where SHAP plots are used to further enhance the user's perception of the data story. Sections one and two are executed by converting the scripts into prompts, leveraging COT techniques with LLMs for generation, while section three directly utilizes the visual outputs generated by SHAP. We demonstrate the script template design of our ABT data stories as follows.

(1) And: the key information in this part includes the model's name, the prediction of the target variable, and the user's question.

- The model's name: the name of the black-box model. It is only the name shown to the user, which does not necessarily contain detailed information about the prediction model itself. For example, although we train XGBoost to predict outputs, the "model name" here can be "BlackBox".
- The prediction: the model's prediction results on the sample inputs. In the Boston Housing dataset, the target variable of prediction is "medv", the median value of owner-occupied homes in USD 1,000's.
- The user's question: The user is given a choice between a "what-if" question and a "why-not" question. The automatic generation of the ABT story is based on the user's choice. 1) The "why-not" question: The non-expert user might have doubts about the prediction result: "Why not other results?" Answering the why-not question can help identify key features that explain why the black-box model predicts a particular value. In this case study, for instance, the user may have doubts about the predicted value of $23,890.2 based on the input feature values and his experiences. 2) The "what-if" question: The non-expert user might be curious about what would happen if certain features in the sample are changed: "What if I change the values of [list of features] to [a new list of values]?" For this question, the features and their values are allowed to be selected and modified to express the user's doubts about the model's prediction process. Subsequently, the black-box model generates a new prediction based on the altered feature values, and an explanation for the resulting change in the predicted output is provided. Modifying multiple feature values at the same time is supported. Illustrating the explanation of predicted output change caused by feature value modification can promote the user's understanding of the impact of these features. For example, a user might be interested in how the model's prediction of house price would change if the location of houses is slightly farther from the city center. In this case, "rm" and "dis" are expected to be higher.

(2) But: In this part, we introduce the iML method to interpret the model's prediction and answer the "why-not" and "what-if" questions. In this case study, we use SHAP to provide local interpretation for specific user questions.

- The "why-not" question: To answer the "why-not" question, we explain the model's prediction by ranking the feature contributions of the sample input. We present SHAP's waterfall plot to visualize feature names, feature contributions, and model output.
- The "what-if" question: To answer the "what-if" question, first, we generate a new prediction based on the user's input. Afterwards, we present a comparison of feature contributions corresponding to the initial and new predictions, respectively, visualized by SHAP's force plots.

(3) Therefore: In this part, a conclusive summary of the explanation is given. Users are expected to draw their judgments about the black-box model's trustworthiness based on our explanation.

- The "why-not" question: The top 3 features with their contributions and contribution percentages are provided. The user is informed of the most influential features in the prediction process.
- The "what-if" question: We compute the change in contributions of features brought about by modifying the feature values. Next, we present the top 3 features according to the contribution changes, which explain how such feature value modifications affect model predictions.

The script above is then rewritten as prompts to feed into LLMs. We provided strict instructions requiring the LLM to generate structured ABT data stories following the provided script. Moreover, we relax the restrictions in the conclusion of the "Therefore" part and require the LLM to output more flexible analytical conclusions based on the context provided in the "And" and "But" parts. We used three LLMs (listed in Section 5.1.4) to generate data story text. Since the three generated texts are similar, we only present the one generated by the Qwen 3 non-thinking model. However, our workflow supports any state-of-the-art (SOTA) LLMs for reasoning and inference.

*5.2.2. Visual Narration of ABT data story models*

With the ABT storytelling framework, non-expert users can quickly identify the fundamental factors influencing the model's prediction. The simple cause-and-effect relation is helpful for quick understanding, which makes this approach especially effective in situations where a concise explanation is needed. To limit the granularity of explanations, we only present the user's input data, prediction values, and contribution of related features in our data story, without revealing feature values and model details.

A what-if ABT story and a why-not ABT story are illustrated in Fig. 6 and Fig. 7. Explanations are given for this specific case. For the what-if question, if we increase the input value of "rm" and "dist", the prediction will increase. This is probably because although the location departs from the city center, the house area and rooms increase simultaneously. Because the latter contributes more to the house price, the black-box model predicts the house price with a higher value. For the why-not question, "rm", "lstat", and "ptratio" have the biggest contributions to the prediction, contributing -1.77, 1.02, and 0.75 to the predicted output, and accounting for 30.41%, 17.49%, and 12.88% of the total contribution.

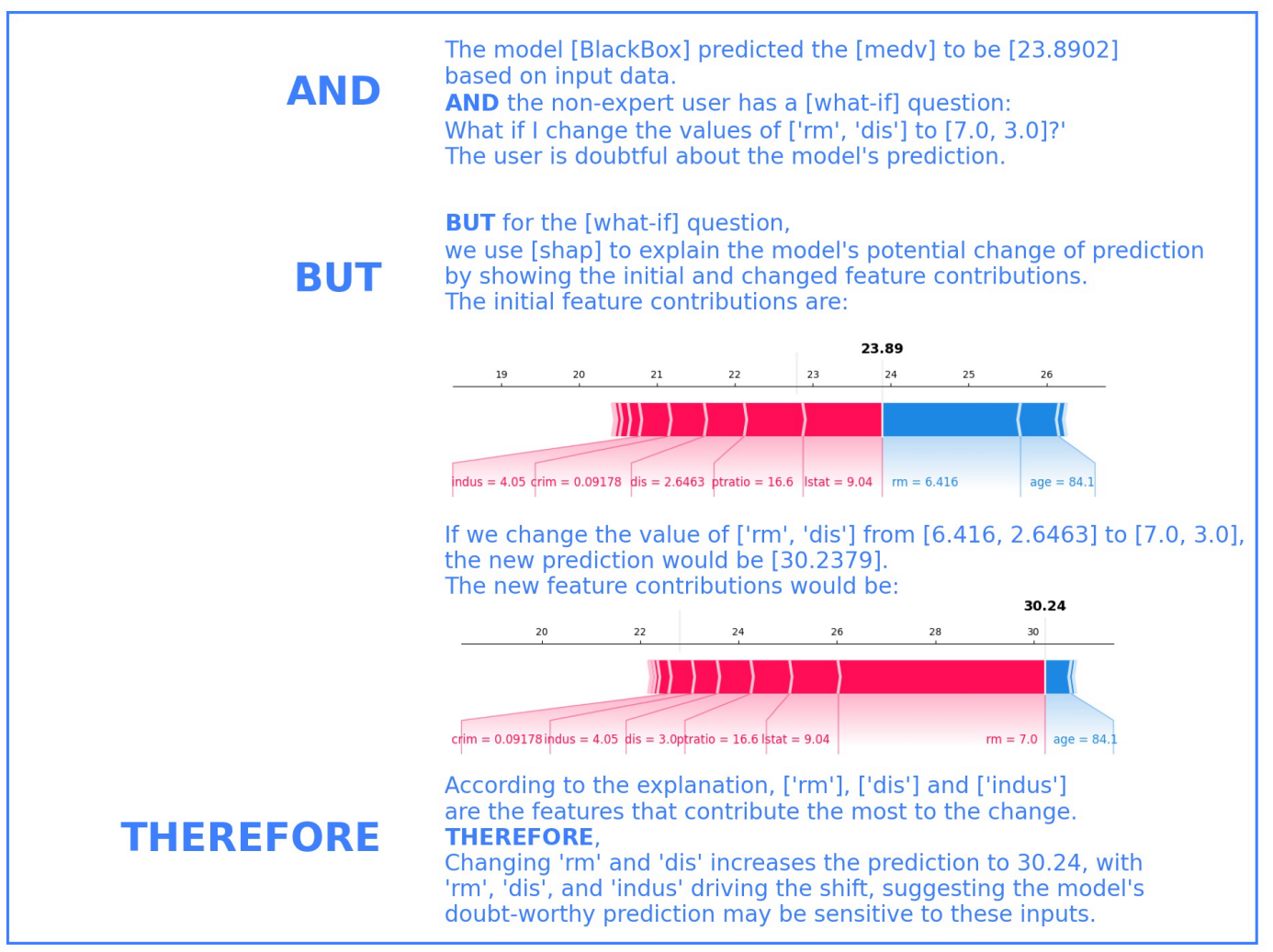

**Fig. 6.** A What-if ABT story[1].

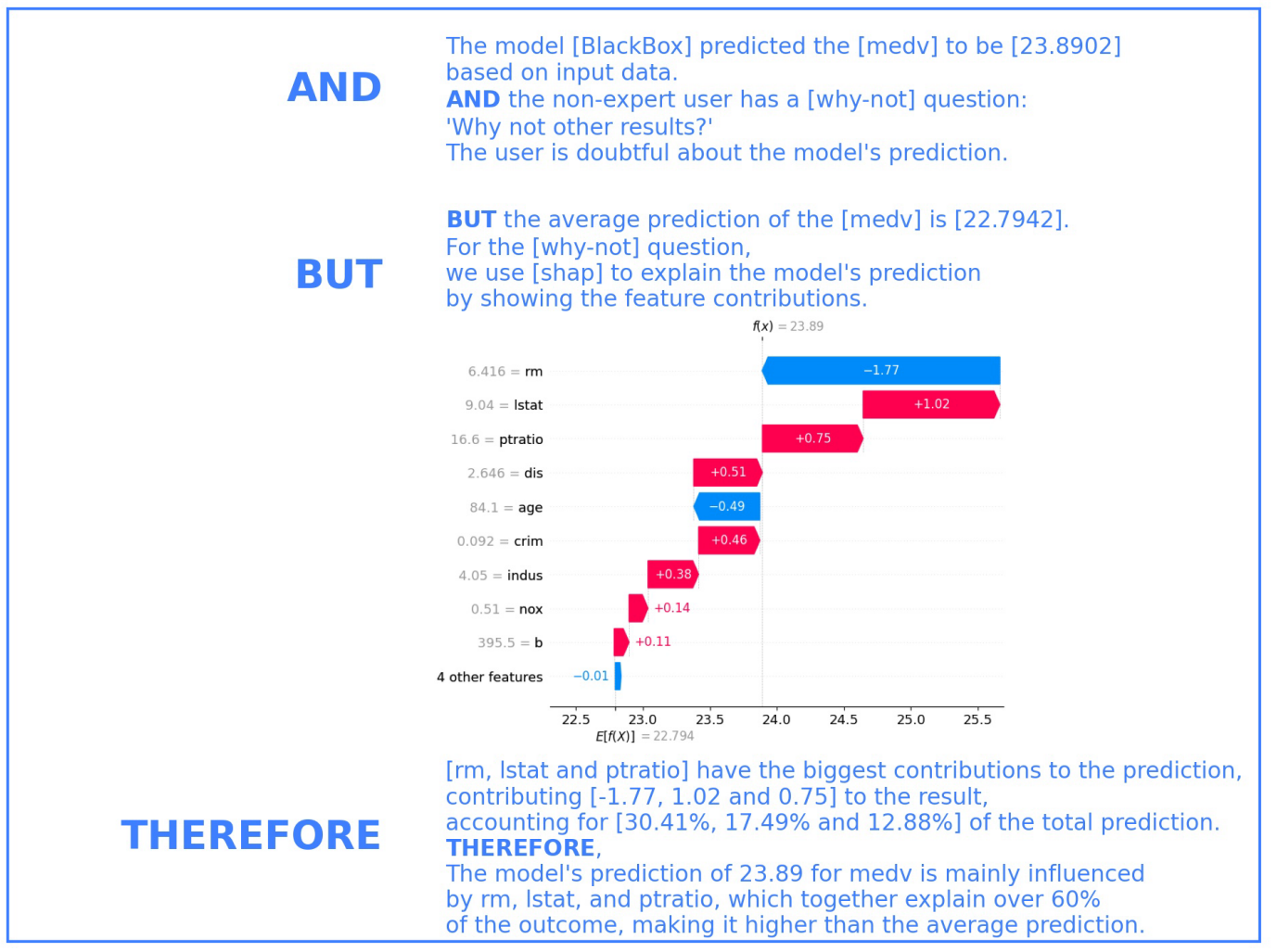

**Fig. 7.** A Why-not ABT story[2].

*5.3. Mapping ABT data stories to non-ABT data stories*

The mapping strategy from ABT data stories to other stories follows the methodology illustrated in Table 3. We coded these predefined rules manually. Four data storytelling frameworks mapping results are demonstrated in Fig. 8 and Fig. 9. The Freytag's Pyramid (Fig. 8.A) and the CAR framework (Fig. 9.A) are used to present the why-not story, while the SPSN framework (Fig. 8.B) and the STAR framework (Fig. 9.B) are designed for the what-if story. This automated mapping process can be extended to other data storytelling frameworks beyond the four examples presented in this section.

[1] Note: This figure is adapted from: https://www.animateyour.science/post/how-to-transform-your-research-into-a-compelling-science-story.
[2] Note: This figure is adapted from: https://www.animateyour.science/post/how-to-transform-your-research-into-a-compelling-science-story.

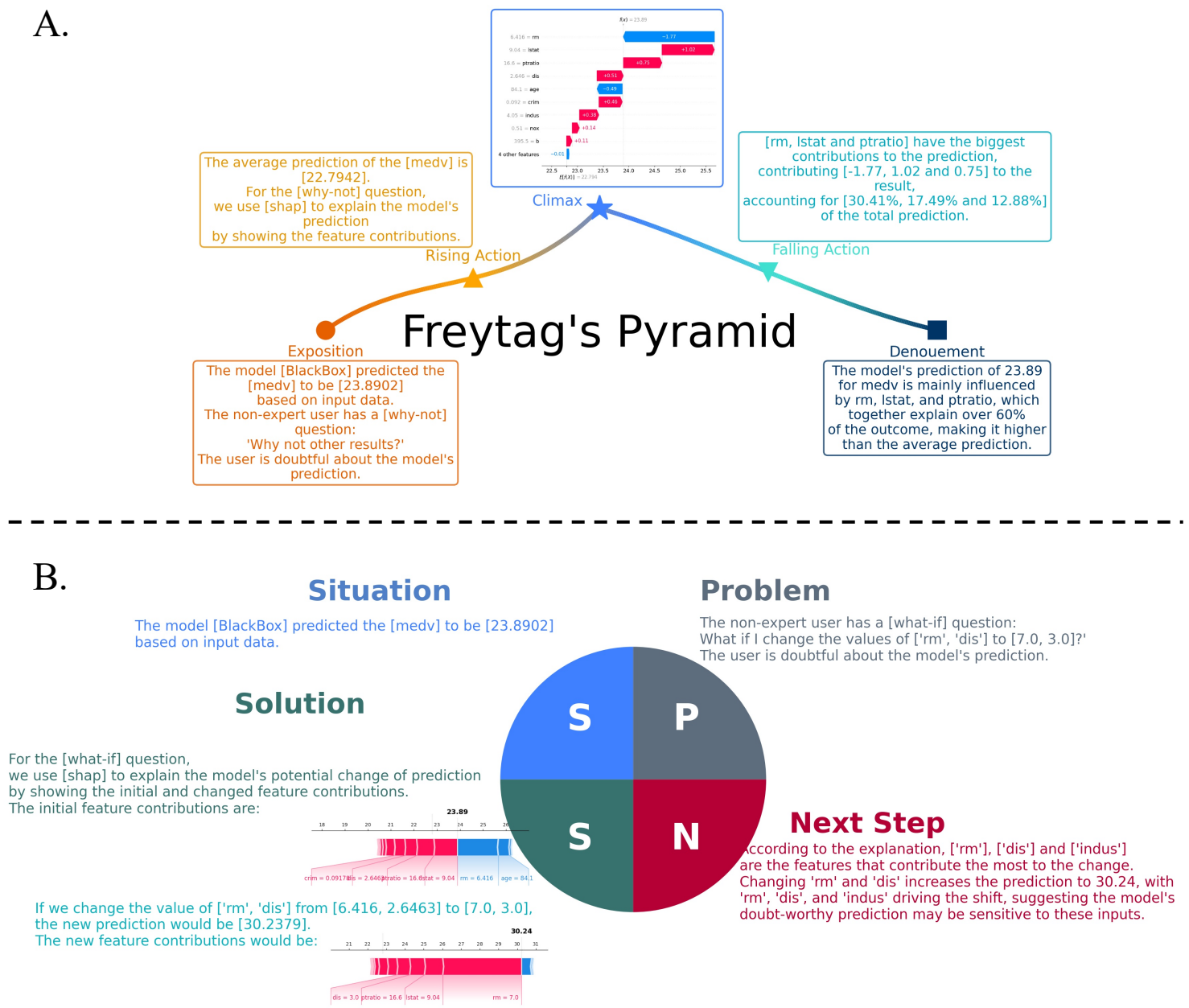

**Fig. 8.** Freytag's Pyramid data story for why-not event (A) and SPSN[3] data story for what-if event (B).

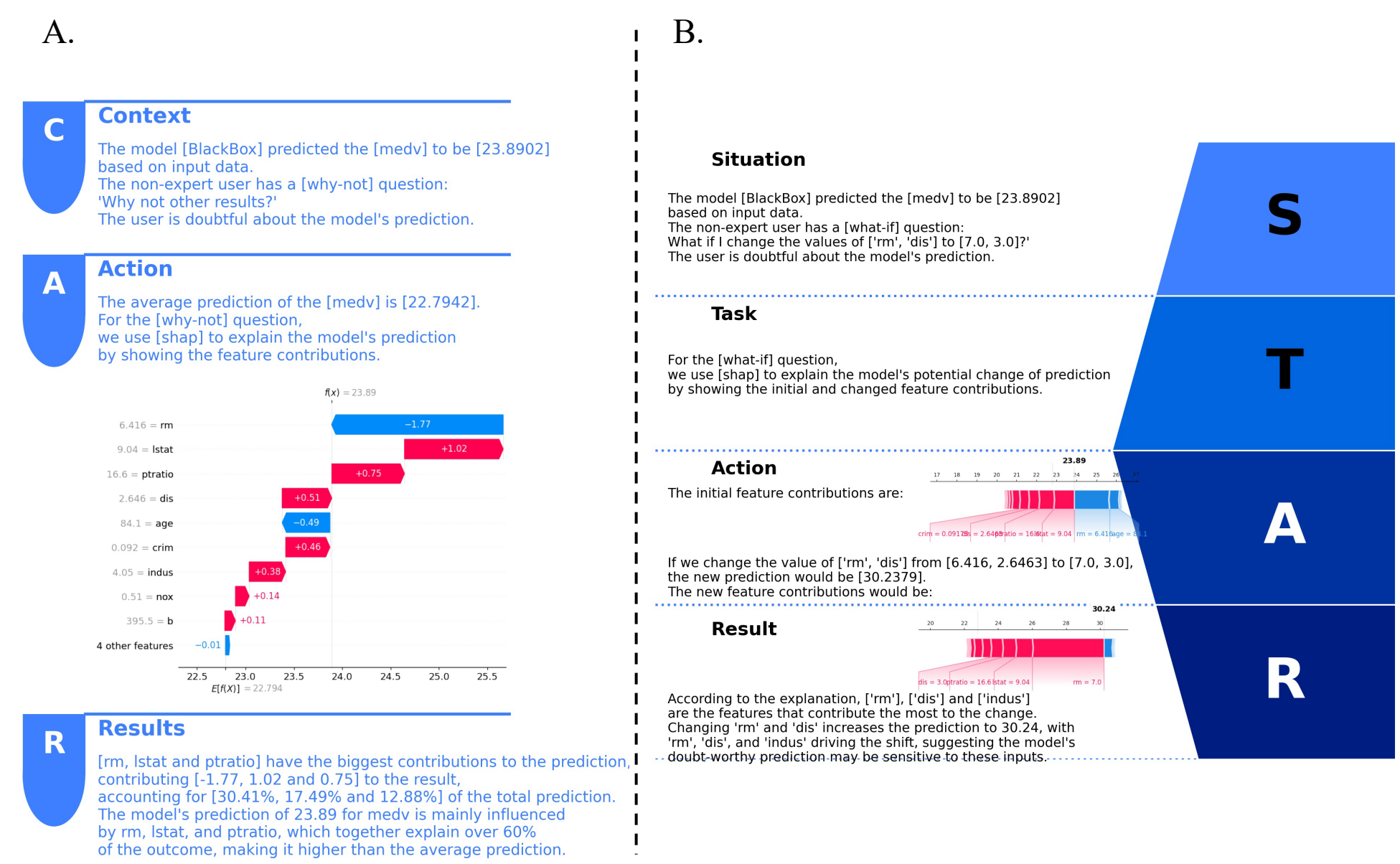

**Fig. 9.** CAR[4] data story for why-not event (A) and STAR[5] data story for what-if event (B).

### *5.4. Evaluation*

To validate the effectiveness of the method we propose, a qualitative analysis is presented to distinguish ABT story explanations with SOTA iML techniques. A questionnaire survey is then conducted to evaluate our ABT data stories quantitatively. This case study uses SHAP plots as a baseline and compares them with our ABT data stories. We generated two sets of materials, both of which include background information, tasks, and dataset descriptions, as well as why-not and what-if events. The difference is that one of the materials uses SHAP visualizations as explanation supports, while the other uses ABT data stories.

[3] Note: This figure is adapted from: https://slidemodel.com/templates/scqa-powerpoint-template/.
[4] Note: This figure is adapted from: https://susannagebauer.com/blog/business-storytelling-frameworks/.
[5] Note: This figure is adapted from: https://www.pipedrive.com/en/blog/aida-model.

*5.4.1 Qualitative analysis*

To evaluate the interpretation quality of iML, researchers employ both quantitative metrics and user studies. Quantitative metrics gauge properties like consistency, fidelity, and stability (Carvalho et al., 2019). However, these quantitative metrics can only be used to measure the performance of the IML technology itself, and do not take into account users' perception and cognition of the explanations it provides. As a result, from the user's perspective, it is difficult to decide which iML technique is the SOTA. Human-centered evaluations, by contrast, involve user studies to assess how explanations affect user trust, understanding, or decision-making (Rong et al., 2024). In this paper, our focus is on the delivery side of explanations, especially for non-experts. Therefore, we conclude seven metrics to support this point of view (Table 5). We also perform a qualitative analysis to state the benefits of our hybrid approach of explaining AI decisions for non-experts.

Compared with iML alone, the hybrid approach of integrating data storytelling with iML techniques offers higher accessibility for non-experts, richer contextualization that aligns explanations with real-world settings, and a coherent narrative structure that enhances logical flow and causal sequencing. These improvements can promote stronger memory retention and better emotional resonance, fostering user trust and engagement. At the same time, the hybrid approach reduces model-dependency and exposure risk, as it can convey key insights without revealing internal architecture or parameters, and it likewise strengthens privacy protection by avoiding disclosure of sensitive data. Collectively, we suppose that integrating data storytelling with iML produces explanations that are more comprehensible, relatable, and secure than those generated by iML techniques in isolation.

**Table 5**

Qualitative comparison of iML techniques with integrated approaches of integrating data storytelling with iML techniques.

| Dimension | Description | iML techniques | iML + data storytelling (This study) |
|---|---|---|---|
| Accessibility to non-experts | The extent to which non-technical audiences can comprehend and engage with AI decisions | Low | High |
| Contextualization | The degree to which explanations reflect real-world settings and align with the audience's knowledge level | Low | High |
| Narrative Structure | Presence of a coherent story structure in the explanation, including logical flow and causal sequencing | Low | High |
| Memory, Recall, and Retention | Effectiveness of the explanation in facilitating long-term understanding and recall of key concepts | Weak | Strong |
| Emotional Resonance | Capacity of the explanation to evoke affective responses such as empathy, trust, or concern | Weak | Strong |
| Model Dependency and Exposure Risk | Likelihood that the explanation methods rely on and potentially expose internal model details (e.g., architecture, parameters, gradients) | Varies(method-dependent) | Low |
| Privacy Protection and Sensitive Data Confidentiality | The extent to which explanations protect privacy and avoid disclosing sensitive or confidential data. | Low | High |

*5.4.2. Survey design*

In this survey, we first asked respondents to read the background information and data descriptions, and then to analyze the explanatory results generated based on what-if and why-not events. To avoid information load, we divided the why-not and what-if events into two groups for the questionnaire and presented ABT data stories and SHAP plots to respondents in random order. Following each explanatory result, we designed five-point scale questions that corresponded to eight evaluation metrics adapted from Table 5 to better fit this case study. Usefulness is added to represent the user's recognition of explanations for certain data story events.

- **Accessibility**: Measures the degree to which the explanation can be easily understood by the user.
- **Contextualization**: Measures the extent to which the explanation aligns with real-world contexts and common sense.
- **Narrative coherence**: Measures the degree to which the explanation follows a logically structured and coherent narrative.
- **Memory retention**: Measures the extent to which the explanation aids long-term memory and recall.
- **Emotional Resonance**: Measures the capacity of the explanation to evoke emotional engagement.
- **Model protection**: Measures the degree to which the explanation avoids exposing internal details of the predictive model.
- **Privacy confidentiality**: Measures the extent to which the explanation protects sensitive or personal data from being revealed.
- **Usefulness**: Measures the degree to which the explanation is useful to answer the user's questions about AI decisions.

This survey is intended for non-experts. Two questionnaires were randomly distributed to 200 respondents each. Questionnaires with shorter response times and those with completely consistent responses were deemed invalid, and the respondents associated with each invalid questionnaire were also considered invalid respondents. After excluding the questionnaires from these invalid respondents, a total of 379 valid questionnaires were obtained, including 188 for why-not events and 191 for what-if events.

*5.4.3. Results*

According to Fig. 10 and Table 6, ABT data stories outperform SHAP plots on accessibility, contextualization, narrative coherence, and usefulness dimensions, exceeding SHAP plots by around 10 percentage points in both why-not and what-if events: 76.1% and 74.3% of respondents agree or strongly agree with the statement that ABT data stories are easy to understand; 84.0% and 84.3% of respondents believe that ABT data stories provide rich contexts and align with their common sense; 80.3% and 81.7% of respondents find that ABT data stories are coherent in narrative structure; 85.6% and 85.3% of respondents consider ABT story as a useful way to answer the user's questions about AI decisions. As for cognitive impact, ABT data stories yield higher memory retention percentages across events, indicating better long-term memory. Although both methods achieve around 50% on emotional resonance metrics, ABT data stories still present higher performance. Lastly, respondents prefer SHAP plots to ABT data stories on model protection and privacy confidentiality, probably because they believe ABT data stories provide more detailed data.

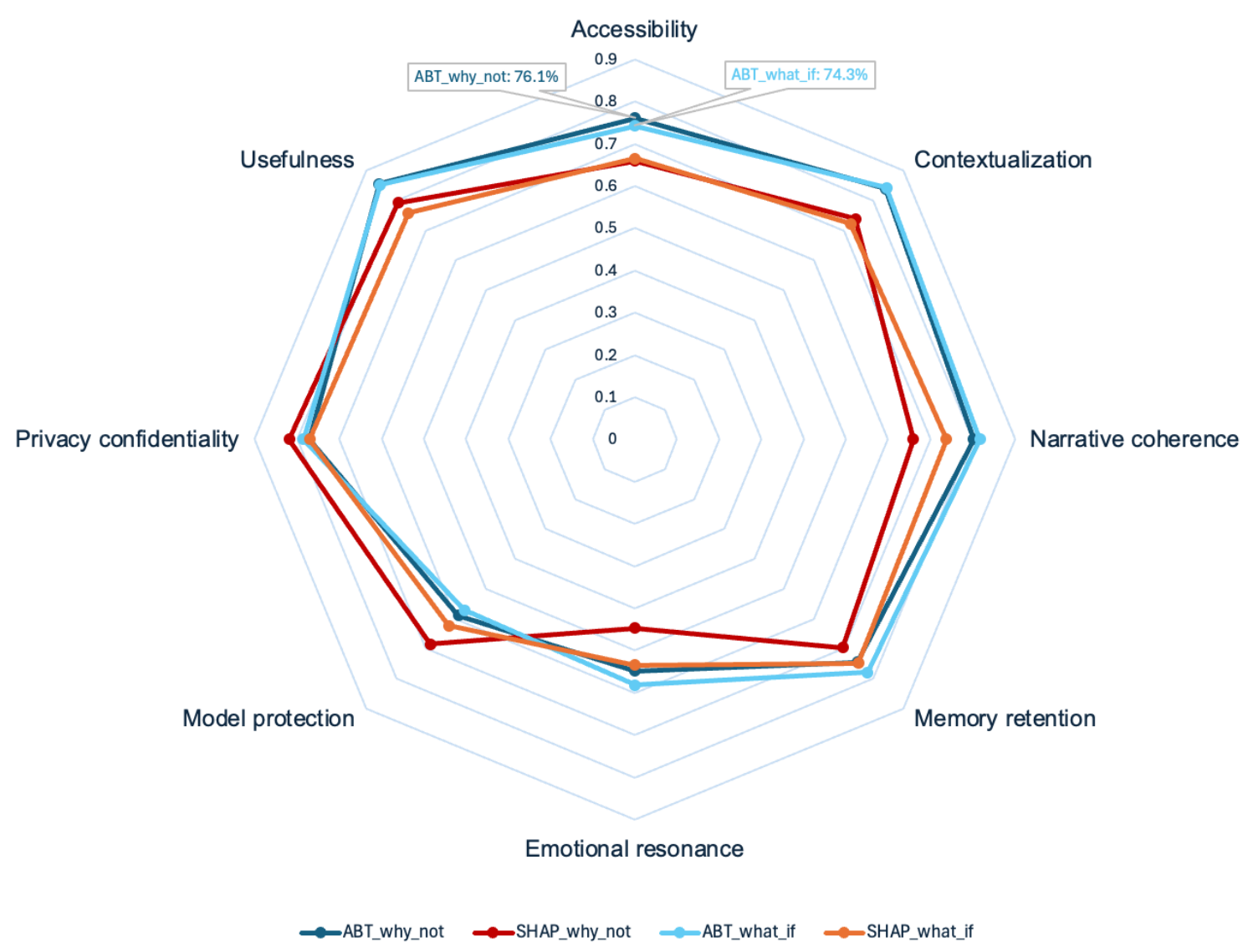

**Fig. 10.** Percentage of respondents selecting "Agree" or "Strongly Agree" of evaluating ABT story and SHAP plot explanations.

The evaluation results in Table 7 indicate the following insights. (1) ABT data stories significantly improve interpretability across key dimensions. Both why-not and what-if events show higher mean scores for accessibility, contextualization, and narrative coherence dimension in ABT data stories compared to SHAP plots, with all differences statistically significant. (2) ABT data stories better support memory and emotional engagement. In Why-not events, ABT data stories lead to significantly higher scores for memory retention and emotional resonance. In What-if events, although improvements in these two dimensions are not statistically significant, mean scores remain higher than SHAP plots. (3) ABT data stories show slightly lower performance in model protection. This may be because the ABT data story provides more detailed information, causing users to be concerned about exposing model risks. However, since the respondents were non-expert users, there might be some limitations in their understanding of the prediction model details. In fact, we intentionally avoided revealing model details when designing our data story structure. (4) No significant difference in privacy protection. The privacy confidentiality scores between ABT data stories and SHAP plots are almost identical in both events, indicating no meaningful impact on users' perceived data sensitivity. (5) The usefulness of ABT data stories is rated higher. The usefulness metric in what-if events shows a significant increase in mean score for ABT data stories. In other words, respondents believe that our ABT data stories are more conducive to answering user questions in what-if events. In summary, despite slightly lower performance in model protection and privacy confidentiality, ABT data stories outperform SHAP plots on core interpretability metrics, including accessibility, contextualization, narrative coherence, memory retention, emotional resonance, and usefulness, yielding superior explanatory performance.

**Table 6**

Percentage of respondents selecting "Agree" or "Strongly Agree" in the evaluation of ABT story and SHAP plot explanations.

| Event types | Metrics | ABT data stories (ours) | SHAP plots |
|---|---|---|---|
| Why-not | Accessibility | **76.1%** | 66.0% |
| | Contextualization | **84.0%** | 73.9% |
| | Narrative coherence | **80.3%** | 66.0% |
| | Memory retention | 74.5% | 69.7% |
| | Emotional resonance | 54.8% | 44.7% |
| | Model protection | 59.0% | **68.6%** |
| | Privacy confidentiality | 77.1% | **81.9%** |
| | Usefulness | 85.6% | 79.3% |
| What-if | Accessibility | **74.3%** | 66.5% |
| | Contextualization | **84.3%** | 72.3% |
| | Narrative coherence | **81.7%** | 73.8% |
| | Memory retention | 78.0% | 74.9% |
| | Emotional resonance | 58.1% | 53.4% |
| | Model protection | 57.1% | **62.3%** |
| | Privacy confidentiality | 78.5% | **77.0%** |
| | Usefulness | **85.3%** | 75.9% |

**Table 7**

Comparison of metric scores for evaluating ABT story and SHAP plot explanations. The number shown in each cell of the third and fourth columns is the mean ± $\sigma$. For each metric, we conduct paired t-tests to assess mean score differences, with statistical significance denoted as follows: ***p < 0.01, **p < 0.05, *p < 0.1.

| Event types | Metrics | ABT data stories (ours) | SHAP plots | Mean difference |
|---|---|---|---|---|
| Why-not | Accessibility | 3.86 ± 0.90 | 3.71 ± 0.93 | **+0.15**** |
| | Contextualization | 4.17 ± 0.86 | 3.96 ± 1.05 | **+0.21***** |
| | Narrative coherence | 4.04 ± 0.89 | 3.79 ± 1.13 | **+0.25***** |
| | Memory retention | 3.87 ± 1.02 | 3.70 ± 1.03 | **+0.17**** |
| | Emotional resonance | 3.48 ± 1.11 | 3.29 ± 1.20 | **+0.21***** |
| | Model protection | 3.55 ± 1.06 | 3.78 ± 1.05 | **−0.23***** |
| | Privacy confidentiality | 3.99 ± 0.94 | 4.01 ± 0.88 | −0.02 |
| | Usefulness | 4.18 ± 0.87 | 4.04 ± 0.97 | +0.14 |
| What-if | Accessibility | 3.87 ± 0.92 | 3.71 ± 0.86 | **+0.16**** |
| | Contextualization | 4.10 ± 0.85 | 3.90 ± 0.95 | **+0.20***** |
| | Narrative coherence | 4.06 ± 0.92 | 3.83 ± 1.03 | **+0.23***** |
| | Memory retention | 4.01 ± 0.95 | 3.90 ± 0.96 | +0.11 |
| | Emotional resonance | 3.57 ± 1.06 | 3.47 ± 1.19 | +0.10 |
| | Model protection | 3.48 ± 1.12 | 3.55 ± 1.11 | −0.07 |
| | Privacy confidentiality | 4.01 ± 0.96 | 4.00 ± 0.92 | +0.01 |
| | Usefulness | 4.20 ± 0.84 | 4.04 ± 0.87 | **+0.16**** |

## 6. Discussion

### *6.1. Theoretical implications*

This study expands data storytelling studies for XAI by introducing an automated method for generating narratives tailored to AI decision-making. By integrating storytelling with AI explainability, the results of black-box models are transformed into comprehensible narratives that are easily accessible to non-experts. Central to this approach are the DIST Pyramid, Formalized Description of Data Storytelling, Data Storytelling Interactive Process, and Data Storytelling Reference Framework, which define the construction and interaction of AI-generated stories. These frameworks build solid theoretical and methodological foundations that have not been seen in existing works, improving the interpretability and trustworthiness of AI decisions, and automating the generation of data stories for non-expert audiences.

This study also advances iML methods by introducing a series of novel frameworks that integrate iML techniques with data storytelling. This integration reduces the complexity of iML technical interpretation, thereby enhancing accessibility to non-experts. The proposed approach shifts the focus from purely technical interpretability to communicative clarity, enhancing trustworthiness and fostering broader adoption of AI systems. It introduces novel research directions in iML, particularly regarding how story structures can effectively articulate the interpretation of AI decisions to diverse audiences.

Based on these contributions, we further propose a Narrative Interpretation Theoretical Framework (Fig. 11), a theoretical framework that integrates iML with data storytelling. This approach enhances the interpretability of machine learning models by embedding explanations within structured narratives. The framework leverages well-established iML techniques, such as LIME, Anchors, and SHAP, to provide insights into model decision-making processes. These methods offer varying levels of interpretability, ranging from local approximations (LIME and Anchors) to global explanations (SHAP). In the context of AI decision storytelling, narrative interpretation can be categorized into two types: local interpretation and global interpretation.

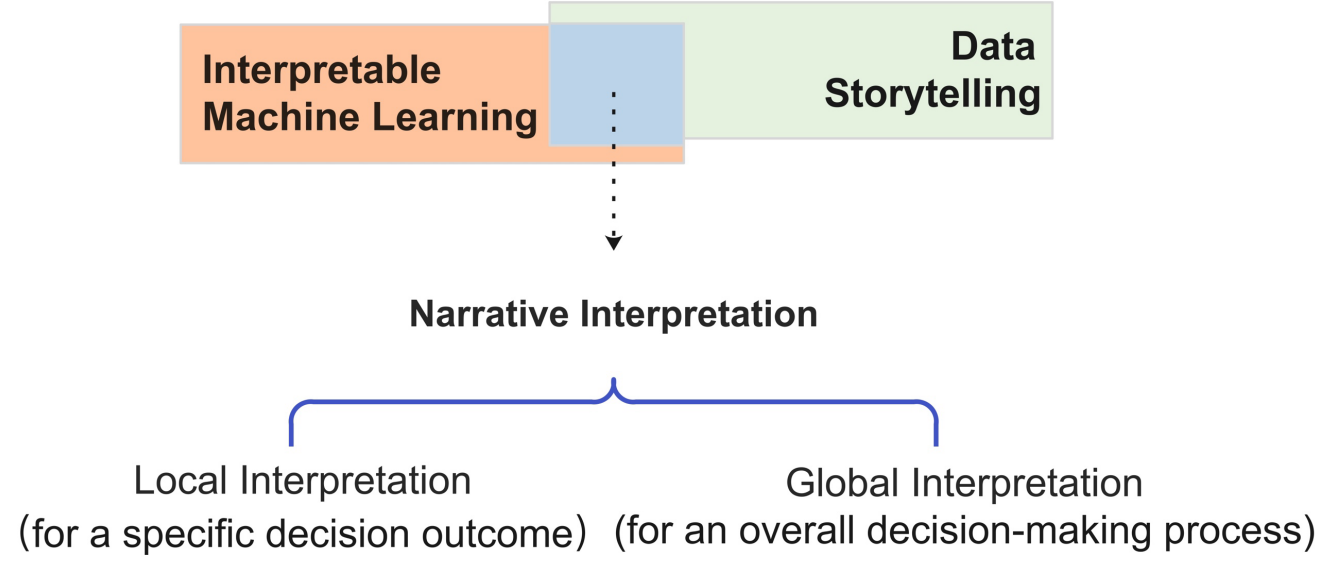

**Fig. 11.** Narrative Interpretation Theoretical Framework.

- Local interpretation pertains to explaining the rationale behind a specific decision made by a machine learning model for an individual input. Techniques such as LIME and SHAP are commonly employed to identify the key features or factors influencing that prediction. This type of interpretation aims to provide a transparent, instance-specific explanation, allowing a clearer understanding of the model's behavior for a given decision.
- Global interpretation, on the other hand, seeks to offer a broader understanding of a black-box model by analyzing its performance across a range of inputs. Methods such as global surrogate models, feature importance analysis, and partial dependence plots (PDPs) are utilized to identify general patterns and trends of the model's decisions. This approach provides insights into the model's broader decision-making process, offering a systematic view of how the model operates.

Narrative interpretation serves to present these AI decisions in a way that is accessible to non-expert audiences. The goal is to convey the reasoning process underlying the AI's outputs in a comprehensible manner, without leaking sensitive data or proprietary model details, thereby enhancing transparency and building trust in the model's decision-making processes. The framework provides a conceptual approach to addressing key challenges in integrating iML with storytelling techniques. It aims to ensure that AI decisions are both accurate and comprehensible while maintaining their relevance to the user's context. Integrating these iML techniques with data storytelling enables narrative interpretation to present complex machine learning outputs in a more accessible and meaningful way for non-experts. Data storytelling plays a pivotal role in elucidating AI decisions for non-experts, ensuring comprehensibility while safeguarding sensitive data and proprietary model details. This ensures that explanations for AI decisions are both transparent and engaging without compromising privacy and security.

### *6.2. Practical Implications*

The practical value and significance of this study lie in its ability to facilitate the communication of AI decisions to non-experts without leaking sensitive data and proprietary model details. This is particularly crucial in high-stake industry domains such as healthcare, finance, and law, where flawed AI decisions can have serious consequences. This study offers a scalable approach to bridge the gap between complex AI systems and non-technical stakeholders. By incorporating data storytelling, this approach provides a more intuitive and engaging way to explain AI decision-making processes. This approach fosters understanding and trust among non-expert decision-makers while protecting data privacy and model secrets.

Furthermore, this study employs LLMs to implement the automated generation of ABT data stories and provides an automatic switch between ABT data stories and other structures of data storytelling. This workflow enhances the effectiveness of AI model communication by tailoring narrative structures to fit various business contexts, thereby ensuring that explanations remain both comprehensible and contextually relevant to a broad audience. Moreover, the primary objective of our research is to establish theoretical frameworks and an implementation workflow for explaining AI decision outcomes based on the DSR paradigm, with data storytelling as the core technique integrated with iML. We conducted the case study to demonstrate the feasibility and flexibility of the implementation workflow we have designed. Based on the qualitative and quantitative evaluation, our ABT data stories outperform SHAP plots overall. However, several deficits should be noted: (1) The cognitive advantages (memory retention and emotional resonance) of data storytelling are not significant in our presented ABT data stories. This is mainly because our ABT data story scripts are designed manually to strictly follow the narrative structure and the objectivity of data, thereby sacrificing the flexibility and expressiveness of natural language. (2) Model protection and privacy protection are implicit for users. We conduct privacy protection techniques on the training data and restrict the granularity of explanations in data stories. However, the non-expert respondents still exhibit low satisfaction on these two dimensions. This implies a discrepancy between technical processing and user perception. More work needs to be done to increase user trust in the protection of model details and privacy.

### *6.3. Limitations*

This study offers valuable insights into integrating data storytelling with iML. Nevertheless, several limitations of this study should be acknowledged despite its contributions:

- Our proposed frameworks were demonstrated in only one case study, which used SHAP to explain predictions of house prices. While this preliminary evidence suggests the feasibility of automatically generating data stories in this specific context, the broader applicability and scalability of the method to other domains and analytical objectives need to be verified through further experimentation.
- Although we employed the ABT story structure to generate "what-if" and "why-not" data stories, our current frameworks do not incorporate more sophisticated story elements that could enhance narrative coherence and user engagement. In this case, users have limited options for various data stories and iML methods to meet their needs. Moreover, the current design only targets a generic non-expert audience, which limits the personalized explanations tailored for specific groups such as policymakers, business leaders, or end-users.
- To the best of our knowledge, there is no standardized methodology for evaluating the quality, relevance, or interpretability of automatically generated data stories. In the absence of established assessment metrics, it is challenging to systematically evaluate the performance of our approach, compare it against alternative data storytelling techniques, or identify improvements.
- Although we used LLMs to generate narrative text in this study, our proposed method relies on manually constructed scripts to ensure the structure of the data story. With the rapid advancement of LLMs, more related techniques can be used to replace manual work and provide more flexible and diverse narrative content.

## 7. Conclusion and future research

This study represents a pioneering effort in applying data storytelling to foster trustworthy AI, addressing gaps in both theoretical and practical research. Previous studies have focused on interpretable machine learning (iML) algorithms, while this study shifts the research focus to automated story generation integrated with iML methods, particularly for explaining AI decisions to non-experts.

Turning data into compelling narratives is a growing area of research in AI, data science, and machine learning. Although this study contributes to the application of data storytelling for trustworthy AI and XAI, there are several key directions for future research:

- Advanced data storytelling with LLMs. There is considerable scope for research in narrative interpretation, particularly in integrating LLMs with the data storytelling process. This could involve the application of detailed techniques such as prompt engineering, requirements specification, and chain-of-thought reasoning to automate the generation of data narratives. Leveraging

LLMs improves data analysis and enables the generation of contextually relevant and coherent narratives, thereby enhancing the interpretability and clarity of AI decision-making. This further reduces reliance on manual methods, thereby improving efficiency and scalability.

- Interactivity and personalization. Enhancing interactivity and personalization in data storytelling can improve user engagement and comprehension. Future research should focus on enabling interactive narratives that incorporate multiple modalities, allowing users to ask questions and receive insights tailored to their specific needs.
- Ethical considerations. Ethical considerations are indispensable in narrative interpretation within data storytelling. Future research requires attention to issues such as fairness, bias mitigation, and privacy protection in the story generation process.

Exploring these research directions will contribute to the advancement of trustworthy AI, enhancing the explanation of AI decisions to non-experts and fostering greater understanding, trust, and engagement with AI systems across diverse domains.

**Declaration of Competing Interest**

The authors declare that they have no known competing financial interests or personal relationships that could have appeared to influence the work reported in this paper.

**References**

Ali, S., Abuhmed, T., El-Sappagh, S., Muhammad, K., Alonso-Moral, J. M., Confalonieri, R., Guidotti, R, Del Ser, J, Díaz-Rodríguez, N, & Herrera, F. (2023). Explainable Artificial Intelligence (XAI): What we know and what is left to attain Trustworthy Artificial Intelligence. *Information fusion*, *99*, 101805. https://doi.org/10.1016/j.inffus.2023.101805

Awosika, T., Shukla, R. M., & Pranggono, B. (2024). Transparency and privacy: the role of explainable ai and federated learning in financial fraud detection. *IEEE Access*, *12*, 64551-64560. https://doi.org/10.1109/ACCESS.2024.3394528

Bataineh, A. S., Zulkernine, M., Abusitta, A., & Halabi, T. (2024). Detecting Poisoning Attacks in Collaborative IDSs of Vehicular Networks Using XAI and Shapley Value. *Journal on Autonomous Transportation Systems*, *2*(3), 1-21. https://doi.org/10.1145/3696462

Beauxis-Aussalet, E., Behrisch, M., Borgo, R., Chau, D. H., Collins, C., Ebert, D., El-Assady, M., Endert, A.,Keim, D. A., Kohlhammer, J., Oelke, D., Peltonen, J., Riveiro, M., Schreck, T., Strobelt, H., Van Wijk, J. J., & Rhyne, T. M. (2021). The Role of Interactive Visualization in Fostering Trust in AI. *IEEE Computer Graphics and Applications*, *41*(6), 7-12. https://doi.org/10.1109/MCG.2021.3107875

Bibi, N., Courtney, J., & McGuinness, K. (2025). Enhancing Brain Disease Diagnosis with XAI: A Review of Recent Studies. *ACM Transactions on Computing for Healthcare*, *6*(2), 1-35. https://doi.org/10.1145/3709152

Bilal, A., Ebert, D., & Lin, B. (2025). Llms for explainable ai: A comprehensive survey. *arXiv preprint arXiv:2504.00125*. https://doi.org/10.48550/arXiv.2504.00125

Borjigin, C. (2021). Automatic Generation and Engineering Research & Development of Data Stories. *Information and Documentation Services*, *42*(2), 53-62. https://doi.org/10.1145/3453156

Caruana, R., Lou, Y., Gehrke, J., Koch, P., Sturm, M., & Elhadad, N. (2015). Intelligible models for healthcare: Predicting pneumonia risk and hospital 30-day readmission. *Proceedings of the 21th ACM SIGKDD international conference on knowledge discovery and data mining* (pp. 1721-1730). https://doi.org/10.1145/2783258.2788613

Carvalho, D. V., Pereira, E. M., & Cardoso, J. S. (2019). Machine learning interpretability: A survey on methods and metrics. *Electronics*, *8*(8), 832. https://doi.org/10.3390/electronics8080832

Catherine, C. (2021). Data Storytelling: How to Tell a Story With Data. Retrieved March 25, 2025, from https://online.hbs.edu/blog/post/data-storytelling

Chen, H., Chiang, R. H., & Storey, V. C. (2012). Business intelligence and analytics: From big data to big impact. *MIS quarterly*, *36*(4), 1165-1188. https://doi.org/10.2307/41703503

Chen, T., & Guestrin, C. (2016). Xgboost: A scalable tree boosting system. In *Proceedings of the 22nd ACM SIGKDD international conference on knowledge discovery and data mining* (pp. 785-794). https://doi.org/10.1145/2939672.2939785

Clegg, T., Greene, D. M., Beard, N., & Brunson, J. (2020). Data everyday: Data literacy practices in a Division I college sports context. In *Proceedings of the 2020 CHI conference on human factors in computing systems* (pp. 1-13). https://doi.org/10.1145/3313831.3376153

Davenport, T. (2014). 10 Kinds of stories to tell with data. *Harvard Business Review*, 6 May 2014. Retrieved March 25, 2025, from https://hbr.org/2014/05/10-kinds-of-stories-to-tell-with-data

Dubey, A., Yang, Z., & Hattab, G. (2024). AI Readiness in Healthcare through Storytelling XAI. *arXiv preprint arXiv:2410.18725*. https://doi.org/10.48550/arXiv.2410.18725

Dykes, B. (2019). *Effective data storytelling: how to drive change with data, narrative and visuals*. John Wiley and Sons.

Echeverria, V., Martinez-Maldonado, R., Granda, R., Chiluiza, K., Conati, C., & Buckingham Shum, S. (2018). Driving data storytelling from learning design. In *Proceedings of the 8th international conference on learning analytics and knowledge* (pp. 131-140). https://doi.org/10.1145/3170358.3170380

European Commission. (2019). Ethics guidelines for trustworthy AI. Retrieved March 25, 2025, from https://ec.europa.eu/digital-single-market/en/news/ethics-guidelines-trustworthy-ai

European Commission. (2024). EU AI Act. Retrieved March 25, 2025, from https://artificialintelligenceact.eu/ai-act-explorer/

Ferracani, A., Bertini, M., Pala, P., Nannotti, G., Principi, F., & Becchi, G. (2024). Personalized Generative Storytelling with AI-Visual Illustrations for the Promotion of Knowledge in Cultural Heritage Tourism. In *Proceedings of the 6th workshop on the analySis, Understanding and proMotion of heritAge Contents* (pp. 28-32). https://doi.org/10.1145/3689094.3689465

Field, S. (2005). *Screenplay: The foundations of screenwriting*. Delta.

Finkler, W., & León, B. (2019). The power of storytelling and video: a visual rhetoric for science communication. *JCOM*, *18*(05), A02-2. https://doi.org/10.22323/2.18050202

Gagnon, E., & Caya, O. (2020). Bridging the Gap between Insights and Action: the Role of Analytical Storytelling. In *Proceedings of the 26th Americas Conference on Information Systems*.

Gallo, C. (2019). The art of persuasion hasn't changed in 2000 years. *Harvard Business Review*, 15 July, 2019. Retrieved March 25, 2025, from https://hbr.org/2019/07/the-art-of-persuasion-hasnt-changed-in-2000-years

Gómez Ortega, A., Bourgeois, J., & Kortuem, G. (2023). Personal data comics: A data storytelling approach supporting personal data literacy. In *Proceedings of the XI Latin American Conference on Human Computer Interaction* (pp. 1-8.). https://doi.org/10.1145/3630970.3630982

Google. (2020). Secure Noise Generation. Retrieved March 25, 2025, from https://github.com/google/differential-privacy/blob/main/common_docs/Secure_Noise_Generation.pdf

Górski, Ł., & Ramakrishna, S. (2021). Explainable artificial intelligence, lawyer's perspective. In *Proceedings of the eighteenth international conference on artificial intelligence and law* (pp. 60-68). https://doi.org/10.1145/3462757.3466145

Graeber, T., Roth, C., & Zimmermann, F. (2024). Stories, statistics, and memory. *The Quarterly Journal of Economics*, *139*(4), 2181-2225. https://doi.org/10.1093/qje/qjae020

Hamon, R., Junklewitz, H., Sanchez, I., Malgieri, G., & De Hert, P. (2022). Bridging the gap between AI and explainability in the GDPR: towards trustworthiness-by-design in automated decision-making. *IEEE Computational Intelligence Magazine*, *17*(1), 72-85. 10.1109/MCI.2021.3129960

Harrison Jr, D., & Rubinfeld, D. L. (1978). Hedonic housing prices and the demand for clean air. *Journal of environmental economics and management*, *5*(1), 81-102. https://doi.org/10.1016/0095-0696(78)90006-2

Hartmann, M., Du, H., Feldhus, N., Kruijff-Korbayová, I., & Sonntag, D. (2022). XAINES: Explaining AI with narratives. *KI-Künstliche Intelligenz*, *36*(3), 287-296. https://doi.org/10.1007/s13218-022-00780-8

Hassan, S., Nadzim, S. Z. A., & Shiratuddin, N. (2015). Strategic use of social media for small business based on the AIDA model. *Procedia-Social and Behavioral Sciences*, *172*, 262-269. https://doi.org/10.1016/j.sbspro.2015.01.363

Heath, C., & Heath, D. (2007). *Made to stick: Why some ideas survive and others die*. Random House.

Hevner, A. R., March, S. T., Park, J., & Ram, S. (2004). Design science in information systems research. *MIS quarterly*, 75-105. https://doi.org/10.2307/25148625

Hoffman, R. R., Lee, J. D., Woods, D. D., Shadbolt, N., Miller, J., & Bradshaw, J. M. (2009). The dynamics of trust in cyberdomains. *IEEE Intelligent Systems*, *24*(6), 5-11. https://doi.org/10.1109/MIS.2009.124

Hoffman, R., Mueller, S., Klein, G., & Litman, J. (2018a). *Measuring Trust in the XAI Context*. Technical Report, DARPA Explainable AI Program. https://doi.org/10.31234/osf.io/e3kv9

Hoffman, R. R., Mueller, S. T., Klein, G., & Litman, J. (2018b). Metrics for Explainable AI: Challenges and Prospects. https://arxiv.org/pdf/1812.04608

Hudon, A., Demazure, T., Karran, A., Léger, P. M., & Sénécal, S. (2021). Explainable artificial intelligence (XAI): how the visualization of AI predictions affects user cognitive load and confidence. In: Davis, F.D., Riedl, R., vom Brocke, J., Léger, PM., Randolph, A.B., Müller-Putz, G. (eds) Information Systems and Neuroscience. NeuroIS 2021. Lecture Notes in Information Systems and Organisation, vol 52. Springer, Cham. https://doi.org/10.1007/978-3-030-88900-5_27

Jain, R. P., Satriadi, K. A., Drogemuller, A., Smith, R., & Cunningham, A. (2024). Once Upon a Data Story: A Preliminary Design Space for Immersive Data Storytelling. In *Companion Proceedings of the 2024 Conference on Interactive Surfaces and Spaces* (pp. 63-68.). https://doi.org/10.1145/3696762.3698054

Kelleher, C., & Pausch, R. (2007). Using storytelling to motivate programming. *Communications of the ACM*, *50*(7), 58-64. https://doi.org/10.1145/1272516.1272540

Knaflic, C.N. (2015). *Storytelling with data: A data visualization guide for business professionals*. John Wiley & Sons.
Knaflic, C.N. (2019). *Storytelling with Data: Let's Practice*. John Wiley & Sons.
Kosara, R., & Mackinlay, J. (2013). Storytelling: The next step for visualization. *Computer*, *46*(5), 44-50. https://doi.org/10.1109/MC.2013.36
Li, H., Wang, Y., & Qu, H. (2024). Where are we so far? understanding data storytelling tools from the perspective of human-ai collaboration. In *Proceedings of the 2024 CHI Conference on Human Factors in Computing Systems* (pp. 1-19). https://doi.org/10.1145/3613904.3642726
Li, Y., Wang, Y., Lee, Y., Chen, H., Petri, A. N., & Cha, T. (2023). Teaching data science through storytelling: Improving undergraduate data literacy. *Thinking Skills and Creativity*, *48*, 101311. https://doi.org/10.1016/j.tsc.2023.101311
Lipton, Z. C. (2018). The mythos of model interpretability: In machine learning, the concept of interpretability is both important and slippery. *Queue*, *16*(3), 31-57. https://doi.org/10.1145/3236386.3241340
Lundberg, S. M., & Lee, S. I. (2017). A unified approach to interpreting model predictions. In *Proceedings of the 31st International Conference on Neural Information Processing Systems* (pp. 4768-4777). https://dl.acm.org/doi/pdf/10.5555/3295222.3295230
Mahoney, J. T., & Nickerson, J. (2022). Oliver Williamson: A hero's journey on the merits. *Journal of Institutional Economics*, *18*(2), 195-207. https://doi.org/10.1017/S1744137421000151
Martens, D., Hinns, J., Dams, C., Vergouwen, M., & Evgeniou, T. (2025). Tell me a story! Narrative-driven XAI with Large Language Models. *Decision Support Systems*, 114402. https://doi.org/10.1016/j.dss.2025.114402
Martinez-Maldonado, R. (2023). Data Storytelling: Revolutionising Human-Data Interaction or Just Passing Hype?. In *Proceedings of the XI Latin American Conference on Human Computer Interaction* (pp. 1-2). https://doi.org/10.1145/3630970.3631007
Mellish, C., Scott, D., Cahill, L., Paiva, D., Evans, R., & Reape, M. (2006). A reference architecture for natural language generation systems. *Natural language engineering*, *12*(1), 1-34. https://doi.org/10.1017/S1351324906004104
McDowell, A. (2019). Storytelling shapes the future. *Journal of Futures Studies*, *23*(3), 105-112. https://doi.org/10.6531/JFS.201903_23(3).0009
Miller, C.H. (2019) *Digital storytelling: A creator's guide to interactive entertainment*. CRC Press.
Minto, B. (1996). *The Minto pyramid principle*. Minto International Inc.
MIT CAPD. (2025). Using the STAR method for your next behavioral interview. Retrieved March 25, 2025, from https://capd.mit.edu/resources/the-star-method-for-behavioral-interviews/.
Olson, R. (2015) *Houston, we have a narrative: Why science needs story*. University of Chicago Press.
Park, K. (2017). *Interactive Storytelling: developing inclusive stories for children and adults*. Routledge.
Ren, D., Brehmer, M., Lee, B., Höllerer, T., & Choe, E. K. (2017). Chartaccent: Annotation for data-driven storytelling. In *2017 IEEE Pacific Visualization Symposium* (pp. 230-239). https://doi.org/10.1109/PACIFICVIS.2017.8031599
Ribeiro, M. T., Singh, S., & Guestrin, C. (2016). "Why should i trust you?" Explaining the predictions of any classifier. In *Proceedings of the 22nd ACM SIGKDD international conference on knowledge discovery and data mining* (pp. 1135-1144). https://doi.org/10.1145/2939672.2939778
Ribeiro, M. T., Singh, S., & Guestrin, C. (2018). Anchors: High-precision model-agnostic explanations. *Proceedings of the AAAI conference on artificial intelligence*, *32*(1). https://doi.org/10.1609/aaai.v32i1.11491
Riche, N. H., Hurter, C., Diakopoulos, N., & Carpendale, S. (Eds.). (2018). *Data-driven storytelling*. CRC Press.
Rong, Y., Leemann, T., Nguyen, T. T., Fiedler, L., Qian, P., Unhelkar, V., ... & Kasneci, E. (2023). Towards human-centered explainable ai: A survey of user studies for model explanations. *IEEE transactions on pattern analysis and machine intelligence*, *46*(4), 2104-2122. https://doi.org/10.1109/TPAMI.2023.3331846
Roth, S. F., Kolojejchick, J., Mattis, J., & Goldstein, J. (1994). Interactive graphic design using automatic presentation knowledge. In *Proceedings of the 1994 CHI conference on Human factors in computing systems* (pp. 112-117). https://doi.org/10.1145/191666.191719
Rudin, C. (2019). Stop explaining black box machine learning models for high stakes decisions and use interpretable models instead. *Nature machine intelligence*, *1*(5), 206-215. https://doi.org/10.1038/s42256-019-0048-x
Ryan, L. (2016). *The visual imperative: Creating a visual culture of data discovery*. Morgan Kaufmann.
Segel, E., & Heer, J. (2010). Narrative visualization: Telling stories with data. *IEEE transactions on visualization and computer graphics*, *16*(6), 1139-1148. https://doi.org/10.1109/TVCG.2010.179
Sinek, S. (2009). *Start with why: How great leaders inspire everyone to take action*. Penguin.
Shao, H., Martinez-Maldonado, R., Echeverria, V., Yan, L., & Gasevic, D. (2024) Data storytelling in data visualisation: Does it enhance the efficiency and effectiveness of information retrieval and insights comprehension?. In *Proceedings of the 2024 CHI Conference on Human Factors in Computing Systems* (pp. 1-21). https://doi.org/10.1145/3613904.3643022

Small, D. A., Loewenstein, G., & Slovic, P. (2007). Sympathy and callousness: The impact of deliberative thought on donations to identifiable and statistical victims. *Organizational Behavior and Human Decision Processes*, *102*(2), 143-153: https://doi.org/10.1016/j.obhdp.2006.01.005

Spodek, J. (2020). Context, Action, Result (CAR): answering interview questions and describing experience effectively. Retrieved March 25, 2025, from https://medium.com/@spodek/context-action-result-car-answering-interview-questions-and-describing-experience-effectively-241f39ce1ff9.

Trim, M., & Butler, E. (2024). Seeing How the Sausage is Made: Data Storytelling as Means and Method in a Computer Science Writing Course. In *Proceedings of the 42nd ACM International Conference on Design of Communication* (pp. 217-222). https://doi.org/10.1145/3641237.3691673

Vom Brocke, J., Hevner, A., & Maedche, A. (2020). Introduction to design science research. *Design science research. Cases*, 1-13. https://doi.org/10.1007/978-3-030-46781-4_1

Vonnegut, K. (2007). *A Man without a Country*. Random House.

Weber, W., Engebretsen, M., & Kennedy, H. (2018). Data stories: Rethinking journalistic storytelling in the context of data journalism. *Studies in communication sciences*, *2018*(1), 191-206. https://doi.org/10.24434/j.scoms.2018.01.013

WebFOCUS Information Center. (2025). Generating Narrative Charts. Retrieved March 25, 2025, from https://webfocusinfocenter.informationbuilders.com/wfappent/TLs/TL_rel/source/charting_narrative.htm.

Wei, J., Wang, X., Schuurmans, D., Bosma, M., Xia, F., Chi, E., ... & Zhou, D. (2022). Chain-of-thought prompting elicits reasoning in large language models. *Advances in neural information processing systems*, *35*, 24824-24837.

Williams, W. R. (2019). Attending to the visual aspects of visual storytelling: using art and design concepts to interpret and compose narratives with images. *Journal of Visual Literacy*, *38*(1-2), 66-82. https://doi.org/10.1080/1051144X.2019.1569832

White, C., Dooley, S., & Roberts, M. (2025). LiveBench: A challenging, contamination-limited LLM benchmark. In *International Conference on Learning Representations (ICLR)*. https://doi.org/10.48550/arXiv.2406.19314

Yang, L., Xu, X., Lan, X., Liu, Z., Guo, S., Shi, Y., Qu, H., & Cao, N. (2021). A design space for applying the freytag's pyramid structure to data stories. *IEEE Transactions on Visualization and Computer Graphics*, *28*(1), 922-932. https://doi.org/10.1109/TVCG.2021.3114774

Zawadzki, J. (2018). Storytelling for Data. Retrieved March 25, 2025, from https://medium.com/data-science/storytelling-for-data-scientists-317c2723aa31

Zdanovic, D., Lembcke, T. J., & Bogers, T. (2022). The influence of data storytelling on the ability to recall information. In *Proceedings of the 2022 Conference on Human Information Interaction and Retrieval* (pp. 67-77). https://doi.org/10.1145/3498366.3505755